\documentclass[11pt,letterpaper]{article}

\usepackage{marsrecon-preprint}
\usepackage[dvipsnames]{xcolor}
\usepackage[hidelinks]{hyperref}
\usepackage{amsmath}
\usepackage{amssymb}
\usepackage{booktabs}
\usepackage{caption}
\usepackage{enumitem}
\usepackage{graphicx}
\usepackage{makecell}
\usepackage{siunitx}
\usepackage{subcaption}
\usepackage{tabularx}

\graphicspath{{Figures/image/}{Figures/dataset/text/}}
\setlist[itemize]{leftmargin=1.15em,itemsep=0.05em,topsep=0.1em}
\title{MarsRecon: Self-Supervised and Multimodal Surface Representations for Mars}
\author{%
  \begin{minipage}[t]{0.31\textwidth}\centering
    \normalsize\textbf{Marius~F.~R.~Juston}\textsuperscript{*}\\[0.5em]
    \small Department of Industrial and Enterprise Systems Engineering\\
    The Grainger College of Engineering\\
    University of Illinois Urbana-Champaign\\
    Urbana, IL 61801-3080, USA\\[0.3em]
    \footnotesize\texttt{mjuston2@illinois.edu}
  \end{minipage}%
  \hspace{0.025\textwidth}%
  \begin{minipage}[t]{0.31\textwidth}\centering
    \normalsize\textbf{Akshay Naik}\textsuperscript{*}\\[0.5em]
    \small Department of Electrical and Computer Engineering\\
    The Grainger College of Engineering\\
    University of Illinois Urbana-Champaign\\
    Urbana, IL 61801-3080, USA\\[0.3em]
    \footnotesize\texttt{akshayn3@illinois.edu}
  \end{minipage}%
  \hspace{0.025\textwidth}%
  \begin{minipage}[t]{0.31\textwidth}\centering
    \normalsize\textbf{Jay Mahajan}\textsuperscript{\textdagger}\\[0.5em]
    \small Department of Computer Science\\
    University of Illinois Urbana-Champaign\\
    Urbana, IL 61801-3080, USA\\[0.3em]
    \footnotesize\texttt{jaym2@illinois.edu}
  \end{minipage}\\[0.9em]
  \small\textsuperscript{*}Equal contribution.\quad
  \textsuperscript{\textdagger}Contributed to manuscript revisions.
}
\date{}

\begin{document}
\maketitle

\begin{abstract}

High-resolution orbital imagery offers a rich record of the Martian surface, but sparse geological labels limit supervised representation learning. We present MarsRecon, a geospatially aware pipeline for learning visual and multimodal representations from HiRISE observations of Olympus Mons. The pipeline calibrates NASA Planetary Data System products, extracts valid georeferenced patches, and trains a masked autoencoder on unlabeled imagery. Increasing input resolution and filtering invalid tokens reduced held-out reconstruction loss from \(0.1751\) to \(0.1342\) in the principal Stage A model series. We then freeze the visual encoder and align its features with observation text, coordinates, and local--global image context. The strongest current local-primary model achieves image-to-text recall@10 of \(0.3787\), text-to-image recall@10 of \(0.9161\), and local-to-global recall@10 of \(0.4350\) on the held-out test split. These results establish a working Mars-specific pretraining and retrieval pipeline; further crop-overlap controls and downstream geological evaluations are needed to assess the broader utility of its embeddings.

\end{abstract}

\section{Introduction and Motivation}

Mars exploration increasingly depends on interpreting orbital imagery at scale. HiRISE aboard the Mars Reconnaissance Orbiter provides some of the highest-resolution orbital views of another planet, resolving crater rims, dust mantling, lava-flow textures, aeolian forms, layered terrains, and other geomorphological structure~\cite{mcewen2010hirise,hirise}. However, learning from HiRISE is not a standard image-classification problem. Observations are long orbital strips with irregular support, inconsistent color availability, product-specific map projections, radiometric calibration terms, and nodata regions.

MarsRecon frames this setting as a foundation-representation problem. Rather than depending on dense labels for every geological class, we construct a patch-level geospatial data pipeline and learn visual representations through masked autoencoding~\cite{he2022mae}. This follows geospatial foundation-model work such as SatMAE and Scale-MAE, where self-supervised transformer pretraining exploits large unlabeled remote-sensing corpora~\cite{cong2022satmae,reed2023scalemae}. The long-term goal is a Mars surface embedding that supports retrieval, mapping, site analysis, and downstream probes. This Mars planetary data, due to the lack of rich environmental features on Mars compared to Earth, makes Mars more difficult to generate embeddings. While Earth could be represented as a feature rich environment with forests, desert, oceans, urban areas and more, Mars is essentially a planet of deserts and variations of crater, lava-flows, aeolian forms, and other geomorphological structures~\cite{mcewen2010hirise,hirise}.

\section{Related Work}

MarsRecon builds on three research threads. First, HiRISE and PDS provide the scientific data foundation for high-resolution Mars surface analysis~\cite{mcewen2010hirise,pdsmro}. Prior Mars machine-learning work has explored crater detection, landform mapping, and terrain analysis, but these supervised tasks are constrained by sparse labels and heterogeneous terrain~\cite{delatte2019,lee2019,nodjoumi2023,huang2026}. Second, geospatial tooling such as TorchGeo supports reproducible spatial sampling and metadata-aware learning pipelines, but Mars requires custom coordinate, projection, and calibration handling~\cite{stewart2025torchgeo}. Third, ViT, MAE, SatMAE, and Scale-MAE provide the technical basis for learning reusable features from unlabeled image patches~\cite{dosovitskiy2021vit,vaswani2017attention,he2022mae,cong2022satmae,reed2023scalemae}. Stage B also connects to CLIP-style contrastive learning and remote-sensing/location-language alignment models such as RemoteCLIP and GeoCLIP~\cite{radford2021clip,liu2024remoteclip,cepeda2023geoclip}.

\section{Stage 0: HiRISE Data Infrastructure}

The project begins with a data layer that turns raw HiRISE products into valid geospatial training samples. Each observation is read with product-specific projection metadata and transformed into a common Mars IAU 2000 geographic coordinate system. Raw DN values are converted into calibrated I/F reflectance using per-product scaling and offset parameters, while nodata masks are preserved so fill regions do not become artificial signal.

Patch sampling is strip-aware because HiRISE bounds often overstate usable support: orbital strips are narrow, angled, and partially empty. The pipeline tracks valid strip footprints and patch validity before producing patch records containing centroid coordinates, bounds, source metadata, valid-channel information, and nodata support. For Olympus Mons, the project processes approximately 149,921 valid patches across 422 strips, with train/validation/test splits designed to reduce geospatial leakage.

\section{Methodology}

\subsection{Stage A: Masked Autoencoder Visual Pretraining}

Stage A learns a Mars visual backbone without geological labels. A calibrated HiRISE patch is split into image tokens, a subset of valid tokens is masked, and a ViT-based masked autoencoder reconstructs the missing content from visible context~\cite{he2022mae}. The encoder is the reusable component; the lightweight decoder is used only during pretraining.

The implemented Stage A path adapts a SatMAE-style MAE recipe to the Mars pipeline~\cite{cong2022satmae}. Mars-specific handling is essential. The trainer uses valid-mask-aware token selection and reconstruction loss, filters or downweights nodata-heavy tokens, and refines validity masks from image tensors at runtime. This prevents the model from spending capacity on fill pixels and focuses the objective on real surface structure. The strongest main-line model uses \(256 \times 256\) image inputs, patch-token size \(8\), mask ratio \(0.5\), strict valid-token filtering, and a ViT-base MAE backbone.

The \(256/p4\) ablation produced a lower reconstruction loss, but we do not treat it as the backbone result. Its smaller patch tokens make reconstruction easier; many masked targets can be close to constant local color or texture patches, so lower pixel loss is not by itself evidence of a more transferable geological representation. We therefore use the \(256/p8\) strict-validity line as the main Stage A result.

\subsection{Stage B: Image--Text--Location Alignment}

Stage B freezes the Stage A encoder and learns a shared embedding space for image, text, coordinate, and local/global context. The implementation uses frozen Stage A image features, frozen T5 text embeddings, coordinate encoders, projection heads, and contrastive losses. A false-negative-aware loss is important because the Olympus split has only 244 unique rationale strings, causing many image patches to share the same text target.

The Stage B experiments proceed from simpler controls to harder spatial alignment. B0 validates frozen-image and frozen-text caches, W\&B/evaluation plumbing, and false-negative-aware retrieval. B1a-geo aligns image, text, and coordinate context. Paired local/global experiments compare an easy center-crop proxy against true expanded geographic crops. The strongest current formulation is local-primary expanded training: the local \(0.005^\circ\) patch remains the main image/text target, while a larger crop is auxiliary global context. We frame Stage B as substantial progress rather than a closed result; the remaining question is which gains reflect learned cross-scale alignment rather than easier crop geometry.

\section{Experimental Setup}

All experiments use the Olympus Mons HiRISE subset generated by the Stage 0 pipeline. Stage A uses held-out reconstruction loss over valid masked support as the primary quantitative metric, supported by qualitative reconstruction previews. Stage B is evaluated as retrieval: image-to-text and text-to-image recall@\(K\), MRR-based alignment score, and local/global retrieval for paired-crop runs.

\section{Results}

\subsection{Stage A Results}

Stage A moved from setup validation to a reproducible MAE training recipe. Table~\ref{tab:stage_a_results} summarizes the main ablation line: validation loss decreased from \(0.1751\) for the \(128/p8\) baseline to \(0.1479\) for \(256/p8\) with valid-mask-aware training, and then to \(0.1342\) with stricter token validity.

\begin{table}[!htbp]
\centering
\caption{\textbf{Stage A masked-autoencoder ablations.} Within the main \(p8\) line, higher resolution and stricter valid-token filtering improve held-out reconstruction loss. The \(256/p4\) row is shown as a caveated ablation because small patch tokens can make reconstruction easier without proving better downstream representation quality.}
\label{tab:stage_a_results}
\small
\begin{tabularx}{\linewidth}{@{}lccccX@{}}
\toprule
\textbf{Run} & \textbf{Image} & \textbf{Patch px} & \textbf{Epochs} & \textbf{Best val loss} & \textbf{Interpretation} \\
\midrule
128/p8 e20 & 128 & 8 & 20 & 0.1751 & Baseline p8 configuration \\
128/p8 e80 & 128 & 8 & 80 & 0.1510 & Longer training helped but remained above 256/p8 \\
256/p8 validmask v1 & 256 & 8 & 20 & 0.1479 & Higher resolution plus valid-mask-aware training \\
256/p8 strict80 & 256 & 8 & 20 & 0.1342 & Best p8 metrics-of-record \\
256/p4 strict80 & 256 & 4 & 20 & 0.0964 & Lower loss but caveated by easier tiny-token reconstruction \\
\bottomrule
\end{tabularx}
\end{table}

These results are best read as an engineering ablation rather than a leaderboard. The p8 improvements isolate two practical lessons: larger inputs expose more geomorphic context, and strict valid-token handling prevents nodata from becoming an easy reconstruction target. The p4 row is retained for transparency, but Stage B uses the p8 backbone to favor transferable context over tiny-patch interpolation.

This choice matters because Stage B freezes the image tower. If the Stage A backbone over-optimizes a pixel task that is too local, the downstream alignment problem inherits a weaker visual representation even when reconstruction loss appears better. The selected p8 backbone is therefore a conservative research choice: it preserves enough spatial context for retrieval while still showing reliable masked reconstruction.

\begin{figure}[!htbp]
\centering
\includegraphics[height=0.29\textheight]{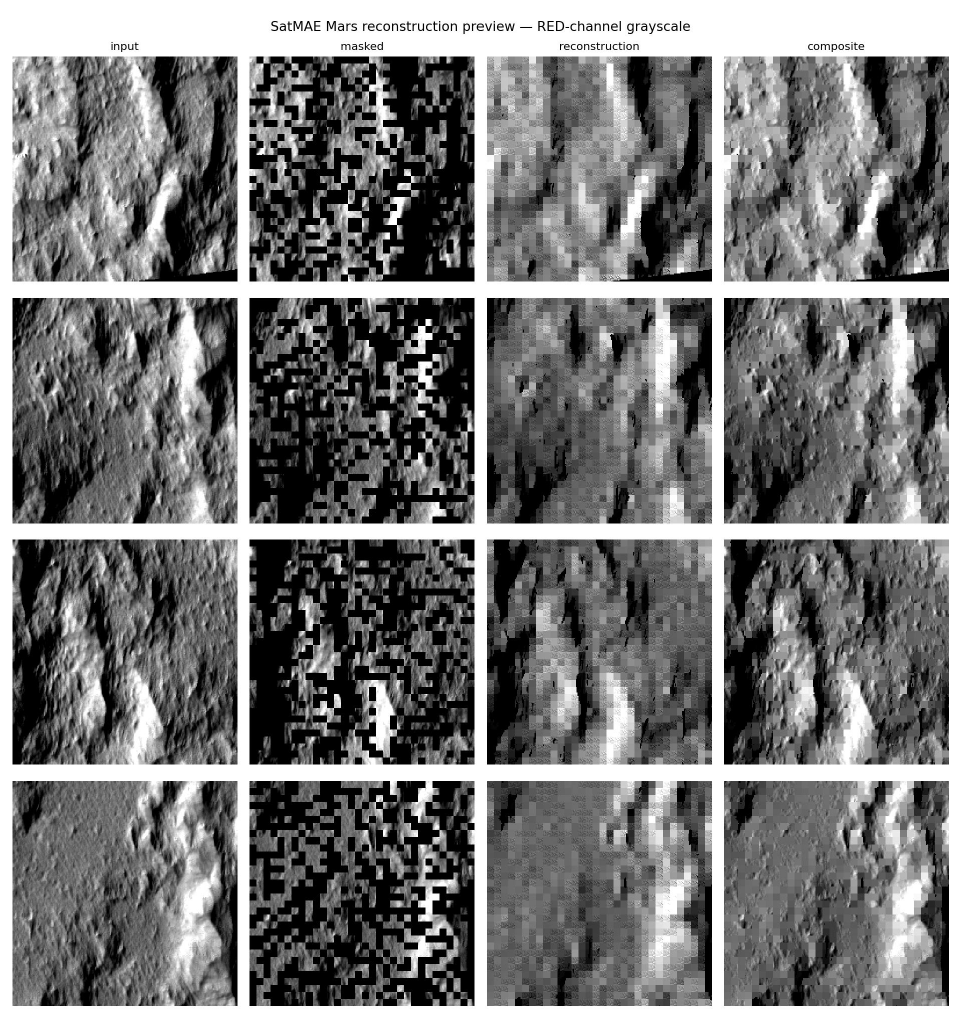}
\caption{\textbf{Stage A reconstruction preview.} Visible context, masked input, reconstruction, and composite preview show that the MAE learns coherent Martian surface structure from valid unlabeled patches.}
\label{fig:stage_a_reconstruction}
\end{figure}

\subsection{Stage B Results}

Stage B results are promising but deliberately framed as in-progress research. The important result is not a single final number; it is that the complete alignment stack now trains, evaluates, and exposes the right failure modes. B0+ showed that frozen Stage A image features can align with text rationales, reaching alignment score \(0.5080\) on full validation. Table~\ref{tab:stage_b_results} shows the higher-signal B1a test runs. B1a-geo+ establishes meaningful image/text/coordinate alignment on the full 12,285-sample held-out test split. The center-crop proxy is strongest but intentionally easier; true expanded \(0.015^\circ\) crops are harder; and local-primary \(0.010^\circ\) is currently the strongest practical candidate.

\begin{table}[!htbp]
\centering
\caption{\textbf{Stage B held-out retrieval summary.} Center-crop retrieval is strong but too easy. True expanded global context is harder. Local-primary training with smaller \(0.010^\circ\) global context gives the most promising current local/global retrieval.}
\label{tab:stage_b_results}
\small
\resizebox{\linewidth}{!}{
\begin{tabular}{lccccp{5.0cm}}
\toprule
\textbf{Run} & \textbf{Align} & \textbf{I2T@10} & \textbf{T2I@10} & \textbf{L2G@10} & \textbf{Interpretation} \\
\midrule
B1a geo+ & 0.4918 & 0.3671 & 0.9068 & n/a & Image/text/coordinate alignment baseline \\
Center proxy & 0.5090 & 0.3867 & 0.9228 & 0.6189 & Strong but easier nested-crop control \\
True expanded 0.015 & 0.4451 & 0.3129 & 0.9057 & 0.2402 & Roadmap-aligned but difficult \\
Local-primary 0.015 lg0.25 & 0.4995 & 0.3744 & 0.8939 & 0.2121 & Recovers image/text quality but weak local/global retrieval \\
Local-primary 0.010 lg0.25 final & 0.4981 & 0.3787 & 0.9161 & 0.4350 & Best practical candidate so far \\
\bottomrule
\end{tabular}
}
\end{table}

The main Stage B conclusion is not that the multimodal model is solved, but that the direction has narrowed. The system has moved from a proposed extension to a working pipeline with interpretable controls. Smaller global contexts improve local/global retrieval, but final controls are needed to separate learned cross-scale alignment from easier crop overlap.

\section{Discussion}

The strongest completed contribution is the Stage 0 to Stage A path: calibrated geospatial data, valid patch sampling, and self-supervised visual pretraining now operate as one reproducible pipeline. Stage A demonstrates that MAE training is viable on Mars HiRISE patches and that Mars-specific valid-mask handling matters. The main limitation is that reconstruction loss measures the pretraining objective, not every downstream representation property; future work should add qualitative retrieval panels and downstream probes for crater, landform, or terrain tasks.

Stage B extends the backbone toward a multimodal Mars embedding, but the results should be framed as active research. The alignment stack works, yet text supervision is weak because many patches share a small number of rationales, and local/global learning is sensitive to crop scale. The retrieval audit in Appendix~\ref{fig:stage_b_authentic_retrieval_appendix} shows that the model can recover plausible local/global and text neighbors, but it also exposes generic Olympus Mons language as a recurring confound. Stage B has made considerable engineering and empirical progress, but the final research claim should remain careful until the last crop-geometry controls are complete.

The most important Stage B lesson is therefore methodological. The center-crop proxy achieves strong retrieval because it preserves an easier visual relationship between local and global views; true expanded crops are harder because they ask the model to connect a local surface patch to a broader surrounding context. That drop should not be treated as failure. It exposes the central scientific confound in planetary embedding learning: a model can appear to improve either because it learns meaningful cross-scale structure or because the crop construction makes the retrieval task geometrically easier. The current \(0.010^\circ\) local-primary results are promising because they recover much of the local/global signal while retaining a more realistic context relationship, but they should be interpreted as a hypothesis that still requires overlap-matched controls and qualitative geological retrieval panels.

This interpretation also clarifies the role of Stage A in the overall project. Stage A is not merely a preprocessing step for Stage B; it is the reusable visual foundation that makes these controlled multimodal experiments possible. By freezing the p8 SatMAE backbone, the project can isolate whether improvements come from text supervision, coordinate encoding, local/global context, or crop geometry rather than from simultaneously changing the visual representation. This modularity is the strongest research position for MarsRecon: the final claim is not that a complete Mars CLIP model has been solved, but that the project has built a calibrated data pipeline, a validated self-supervised Mars backbone, and an auditable experimental framework for testing whether Mars surface embeddings become scientifically meaningful.

\section{Limitations and Future Work}

The main limitation of the current study is scope. The experiments are centered on Olympus Mons, which is scientifically useful because it contains diverse volcanic, erosional, and aeolian terrain, but it is not a full-Mars benchmark. A stronger future evaluation should expand to additional geographic regions and enforce spatially disjoint splits at the observation or strip level. This would test whether the learned representation generalizes across terrain regimes rather than memorizing local imaging conditions, repeated HiRISE observation rationales, or region-specific textures.

A second limitation is evaluation. Stage A reconstruction loss is appropriate for selecting a stable masked-autoencoder checkpoint, but it is an indirect proxy for geological usefulness. Stage B retrieval metrics are more task-oriented, but they still depend on coarse observation-level rationales and crop construction choices. The qualitative retrieval panel is a first audit of geological coherence; the stronger next step is downstream probing with crater, landform, or terrain classification labels when such labels are available. These tests would connect the learned embedding to scientific utility rather than only to self-supervised losses.

Finally, the current system is strong enough to support a more ambitious next study. The infrastructure now supports calibrated HiRISE patch extraction, valid-mask-aware self-supervised pretraining, frozen-feature caching, multimodal alignment, logged ablations, and reproducible held-out evaluation. Future work can build on this foundation by adding viewing-geometry conditioning, stronger text supervision, larger geographic coverage, and comparisons against generic Earth-trained remote-sensing encoders. The central open question is no longer whether the pipeline can be built; it is which combination of Mars-specific data handling and multimodal supervision produces embeddings that planetary scientists would trust for search, mapping, and discovery.

\section{Conclusion}

MarsRecon demonstrates a practical path toward reusable Mars surface representations from HiRISE imagery. The project first solves the geospatial data problem, then trains a self-supervised Stage A visual backbone on unlabeled Martian patches. The strongest Stage A result reduces p8 validation reconstruction loss to \(0.1342\) through higher-resolution inputs and stricter valid-token filtering. The Stage B experiments further show that this backbone can support image--text--location alignment. Stage B remains in progress, but the current evidence is strong enough to define a focused next step: validate whether the \(0.010^\circ\) local/global gains reflect real cross-scale learning or easier crop geometry.

The broader contribution is the staged research design. MarsRecon separates the planetary representation-learning problem into auditable components: calibrated data construction, valid-mask-aware self-supervised visual learning, and multimodal alignment with explicit controls. This separation is important because Mars data introduces failure modes that are easy to hide in an end-to-end model, including nodata artifacts, repeated observation rationales, regional sampling bias, and crop-overlap shortcuts. By making those issues visible, the project establishes a stronger foundation for future work than a single opaque embedding model would provide.

The evidence therefore supports three bounded claims. First, the Stage 0 pipeline produces enough valid, calibrated HiRISE patches to train modern self-supervised models reproducibly. Second, the Stage A p8 MAE backbone improves reconstruction under stricter Mars-specific validity handling and is a defensible frozen visual representation. Third, Stage B demonstrates measurable image--text--location alignment, but its strongest local/global result should be treated as an active hypothesis until crop-overlap controls and qualitative retrieval analysis confirm geological meaning.

The immediate path forward is therefore clear: keep the Stage A backbone fixed as the visual foundation, complete the remaining Stage B crop-geometry controls, and add qualitative retrieval panels that show whether nearest-neighbor results are geologically meaningful. If those controls hold, MarsRecon becomes a credible starting point for downstream Mars tasks such as crater/landform probing \cite{Robbins2012AParameters, Zhuo2025High-AccuracySurfaces}, site search, and large-scale surface similarity mapping.  Crater detection in particular was not performed due to the low availability of datasets. The human labeled and validated dataset from NASA only supports craters of size greater than 1 km; however, the HiRISE resolution is much smaller than that \cite{Robbins2012AParameters}. The other datasets are generated using machine learning \cite{Zhuo2025High-AccuracySurfaces, DoranMars3.2} and would need thorough validation of the datasets.

\bibliographystyle{unsrt}
\bibliography{references}

\appendix

\section{Custom Mars-Specific MAE Architecture}

Although the final Stage A backbone follows the SatMAE-style path, the project also implemented a Mars-specific MAE comparison model. This was useful for validating which parts of the problem were architectural and which parts were caused by Mars data handling.

\begin{itemize}
    \item \textbf{Purpose.} A Mars-specific model within the ViT/MAE family, used to compare standardized remote-sensing MAE design against Mars-oriented valid-mask handling.
    \item \textbf{Input.} A multispectral HiRISE patch, a valid-pixel mask, and patch-level scale features.
    \item \textbf{Patch tokenization.} Non-overlapping patches are embedded with a convolutional patch projection before the transformer blocks.
    \item \textbf{Encoder.} The encoder processes visible valid tokens only. Small presets used low-dimensional encoders for smoke tests; larger presets followed a ViT-base-style configuration.
    \item \textbf{Decoder.} A lighter decoder reconstructs masked patches after encoder latents are projected into decoder space.
    \item \textbf{Validity-aware masking.} Patch-level valid-pixel fractions are computed from the HiRISE nodata mask. Invalid or nodata-heavy patches are excluded from visible-token selection and masked reconstruction loss.
    \item \textbf{Visible-token packing.} Valid visible tokens are packed for efficient encoding and scattered back into spatial positions before reconstruction.
    \item \textbf{Scale conditioning.} Tokens can receive sinusoidal scale encodings, following the motivation of scale-aware geospatial transformers~\cite{reed2023scalemae}.
    \item \textbf{Loss.} Reconstruction loss is computed only on masked pixels that are also spatially valid, reducing the influence of fill regions.
\end{itemize}

\section{Stage A Additional Notes}

The \(256/p4\) run is intentionally not used as the main Stage A backbone despite its lower reconstruction loss. The smaller token patch can make the objective easier by turning many targets into locally smooth or near-constant reconstruction patches. A fair claim of superiority would require downstream retrieval, probing, or qualitative evidence beyond pixel loss.

\begin{figure}[!htbp]
\centering
\includegraphics[width=0.92\linewidth]{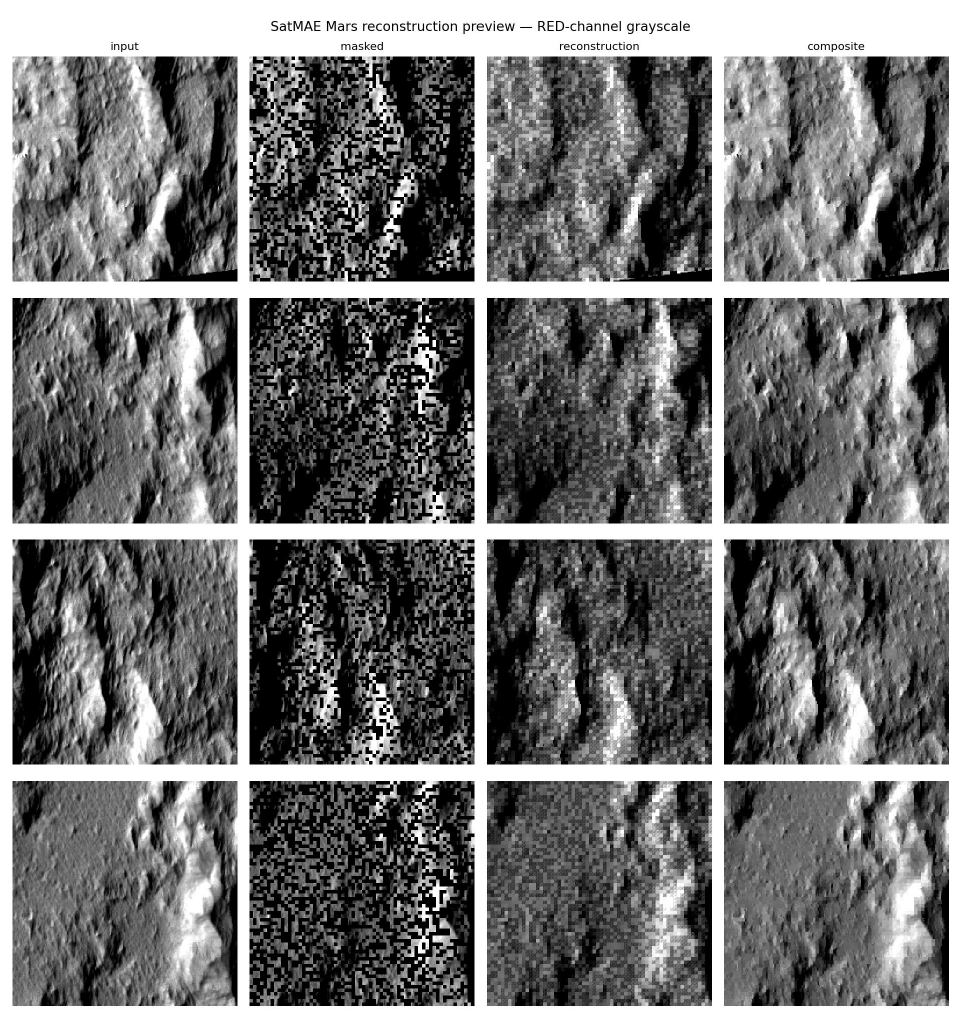}
\caption{\textbf{Caveated \(256/p4\) reconstruction preview.} The lower pixel loss of this run should not be interpreted as a stronger backbone without additional representation-quality evidence, because smaller patch tokens can make the reconstruction target easier.}
\label{fig:p4_caveat}
\end{figure}

\section{Stage B Full Metrics Ledger}

\begin{table}[!htbp]
\centering
\caption{\textbf{Stage B full metrics ledger.} This table preserves the broader run history used to interpret the compact Stage B result table in the main text. Center-crop is a high-capacity/easy control, train-only augmentation is a negative result, and local-primary \(0.010^\circ\) runs are the strongest current practical candidates.}
\label{tab:stage_b_full_ledger}
\scriptsize
\setlength{\tabcolsep}{2.4pt}
\resizebox{\linewidth}{!}{
\begin{tabular}{lrrrrrrrr}
\toprule
\textbf{Run} & \textbf{N} & \textbf{Align} & \textbf{I2T@10} & \textbf{T2I@10} & \textbf{GlobalText} & \textbf{L2G@10} & \textbf{G2L@10} & \textbf{Overlap@10} \\
\midrule
B1a geo latlon & 12285 & 0.4736 & 0.3410 & 0.9289 & n/a & n/a & n/a & n/a \\
B1a geo+ coords rff-siren & 12285 & 0.4918 & 0.3671 & 0.9068 & n/a & n/a & n/a & n/a \\
center-crop proxy & 12285 & 0.5090 & 0.3867 & 0.9228 & n/a & 0.6189 & 0.6170 & n/a \\
true expanded 0.015 & 12285 & 0.4451 & 0.3129 & 0.9057 & n/a & 0.2402 & 0.2470 & n/a \\
train-only augment & 12285 & 0.0820 & 0.0609 & 0.1310 & n/a & 0.2401 & 0.2447 & n/a \\
dual text expanded & 12285 & 0.4463 & 0.2595 & 0.9036 & n/a & 0.2214 & 0.2109 & n/a \\
local-primary 0.015 lg0 & 12285 & 0.4864 & 0.3613 & 0.9280 & 0.4111 & 0.1970 & 0.1696 & 0.4314 \\
local-primary 0.015 lg0.10 & 12285 & 0.4946 & 0.3374 & 0.9196 & 0.4251 & 0.2069 & 0.1853 & n/a \\
local-primary 0.015 lg0.25 sched & 12285 & 0.4995 & 0.3744 & 0.8939 & 0.4344 & 0.2121 & 0.1954 & 0.4519 \\
local-primary 0.010 lg0.25 best & 12285 & 0.4961 & 0.3605 & 0.9141 & 0.4857 & 0.4343 & 0.4353 & 0.5394 \\
local-primary 0.010 lg0.25 final & 12285 & 0.4981 & 0.3787 & 0.9161 & 0.4914 & 0.4350 & 0.4367 & 0.5403 \\
local-primary 0.0125 lg0.25 best & 12285 & 0.5041 & 0.3335 & 0.9133 & 0.4508 & 0.3058 & 0.2902 & 0.4475 \\
local-primary 0.0125 lg0.25 final & 12285 & 0.4982 & 0.3748 & 0.9045 & 0.4636 & 0.3113 & 0.3030 & 0.4564 \\
local-primary 0.0125 lg0.10 & 12285 & 0.4925 & 0.3722 & 0.9088 & 0.4493 & 0.3053 & 0.2930 & 0.4497 \\
local-primary 0.010 lg0.10 best & 12285 & 0.5013 & 0.3552 & 0.9025 & 0.4808 & 0.4268 & 0.4255 & 0.5318 \\
local-primary 0.010 lg0.10 final & 12285 & 0.4915 & 0.3751 & 0.8986 & 0.4893 & 0.4280 & 0.4273 & 0.5343 \\
\bottomrule
\end{tabular}
}
\end{table}

\section{Stage B Design Context}

Stage B aims to turn the Stage A visual backbone into a richer Mars surface embedding. Three non-visual information sources are relevant. First, text rationales describe why a HiRISE observation was collected; this motivates frozen T5-style text embeddings~\cite{t5} and image--text retrieval evaluation. Second, planetary coordinates provide location context, so Stage B includes latitude/longitude encoders rather than treating each patch as an isolated image. Third, orbital viewing geometry such as incidence, emission, and phase angle can affect observed appearance and may be incorporated through FiLM-style conditioning~\cite{Perez2017FiLM:Layer}. The Stage B experiments reported here focus on frozen image/text caches, coordinate encoders, false-negative-aware contrastive alignment, and local/global crop retrieval; viewing-geometry conditioning remains future work.

\section{Pipeline Overview}

\begin{figure}[!htbp]
    \centering
    \includegraphics[width=\linewidth]{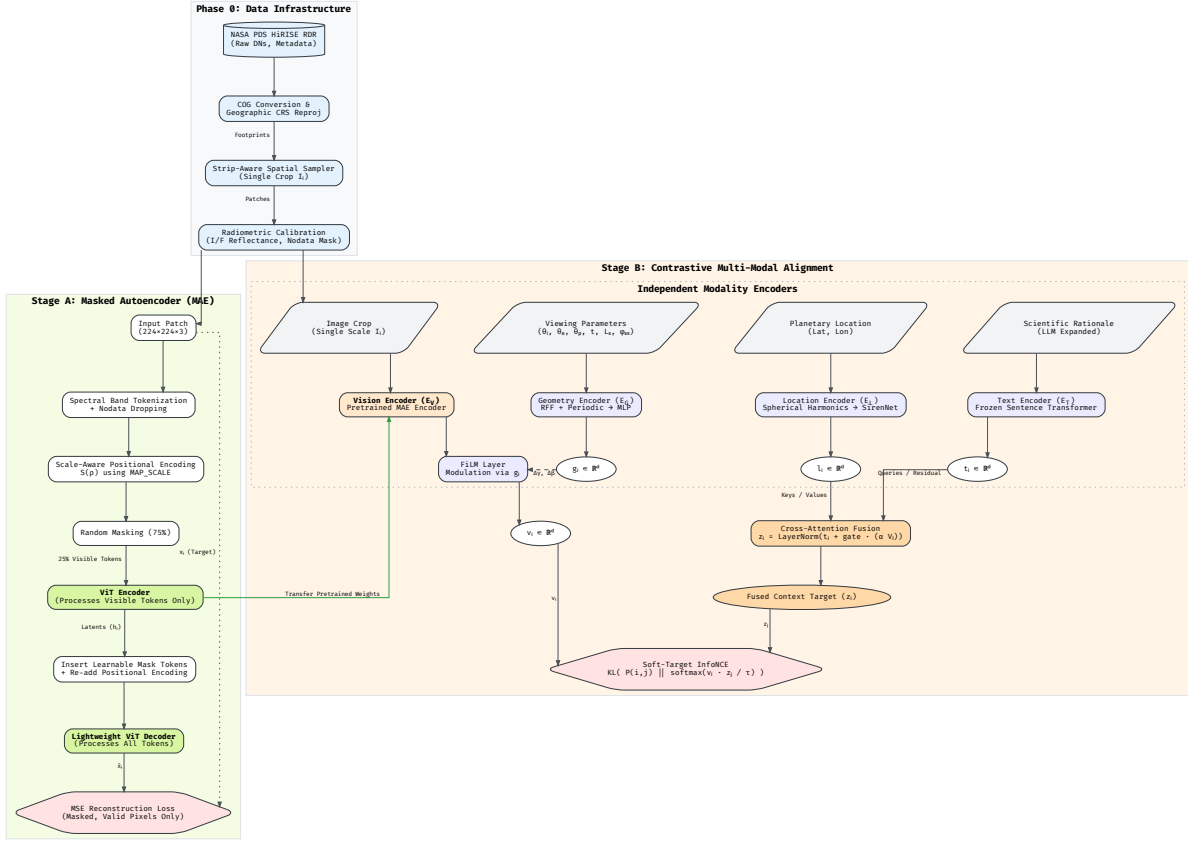}
    \caption{\textbf{MarsRecon pipeline architecture.} Stage 0 ingests NASA PDS HiRISE products, performs reprojection and radiometric calibration, and emits valid patch-level training samples. Stage A trains a masked autoencoder on valid Mars surface tokens. Stage B extends the frozen visual backbone into a multimodal embedding space using text, coordinate, and local/global context.}
    \label{fig:marsclip_architecture}
\end{figure}

Figure~\ref{fig:marsclip_architecture} summarizes the MarsRecon pipeline: Stage 0 builds calibrated geospatial patches, Stage A learns a self-supervised visual backbone, and Stage B aligns the learned visual representation with non-visual context. The Stage B figure illustrates the broader research design, including components that remain under development.

\begin{figure}[!htbp]
    \centering
    \includegraphics[height=0.8\textheight]{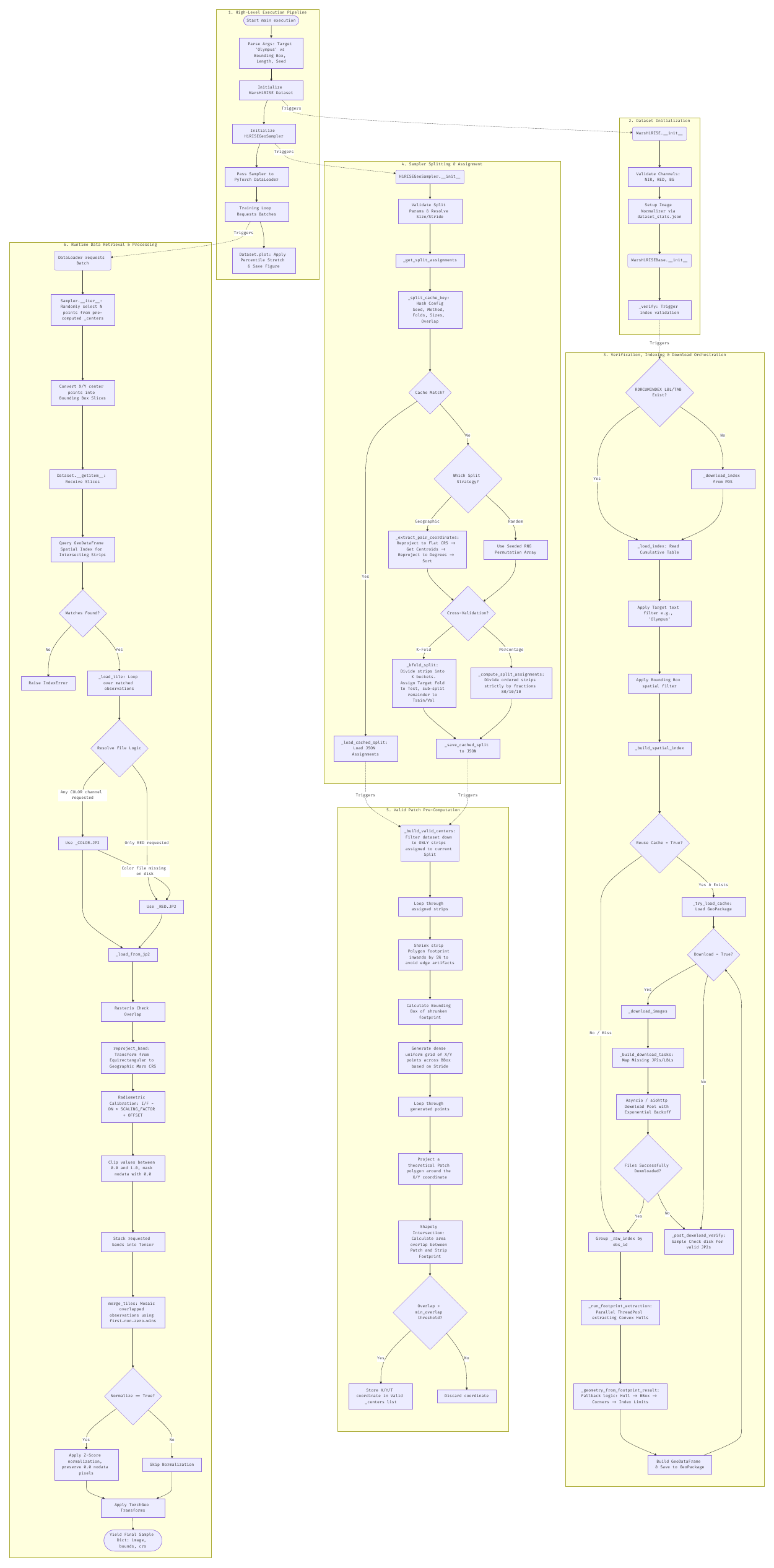}
    \caption{\textbf{Dataset loader and sampler architecture.} The data system connects PDS product discovery, Mars CRS/projection handling, radiometric calibration, valid-footprint estimation, and patch sampling into one trainable HiRISE dataset interface.}
    \label{fig:dataset_architecture_appendix}
\end{figure}

\section{Validation Details}

\subsection{Stage A Validation}

Stage A evaluates whether the ViT-based masked autoencoder learns the distribution of Martian surface textures without labels. The primary metric is masked reconstruction loss over valid pixels, which is appropriate for monitoring optimization but should not be treated as a complete representation-quality metric. Mean squared error can favor smooth reconstructions, so qualitative reconstruction previews and downstream probes remain important complements~\cite{jiang2021ffl,Rahaman2018OnNetworks}.

\subsection{Stage B Validation}

Stage B is evaluated as a retrieval problem in the learned embedding space. Image-to-text and text-to-image recall@\(K\) measure cross-modal alignment, while MRR-based alignment score summarizes rank quality. For paired local/global experiments, local-to-global and global-to-local recall@\(K\) measure whether the model retrieves the corresponding context crop. These metrics are intentionally stricter than training loss because they evaluate whether the embedding is useful for search and association rather than merely optimizing a contrastive objective.

\begin{figure}[!htbp]
    \centering
    \includegraphics[width=0.96\linewidth]{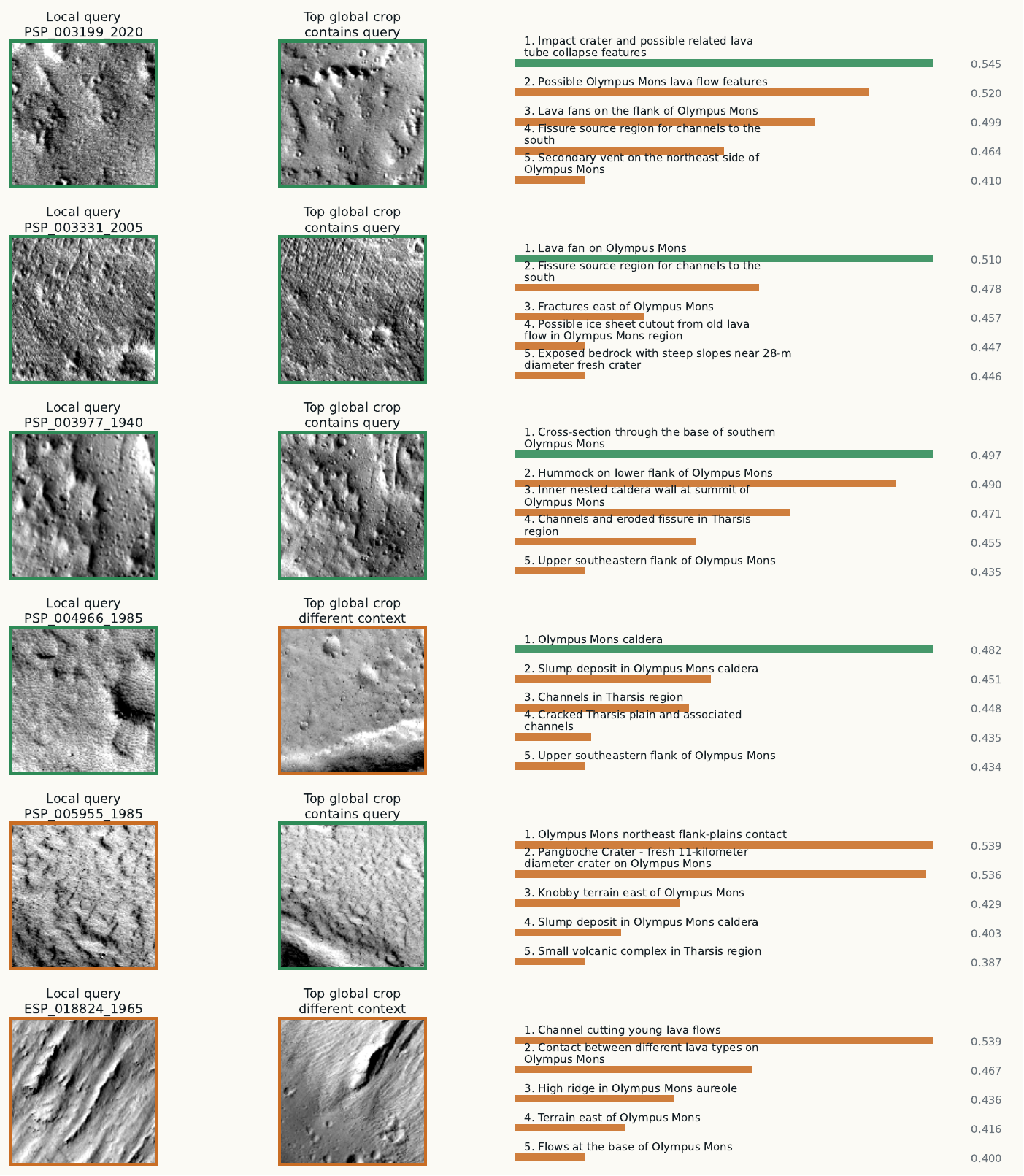}
    \caption{\textbf{Authentic Stage B retrieval examples from the held-out test set.} Each row shows a real local query patch, the model's top retrieved global-context crop, and the top five retrieved rationale strings from the final local-primary \(0.010^\circ\)/\(0.025^\circ\) Stage B run. Green text bars mark the true rationale when it appears in the top five; orange bars show confounding alternatives. Thumbnails are grayscale contrast-stretched for display, but the ranks and similarity scores are computed directly from exported model embeddings.}
    \label{fig:stage_b_authentic_retrieval_appendix}
\end{figure}

Figure~\ref{fig:stage_b_authentic_retrieval_appendix} provides a qualitative audit of the final local-primary Stage B run. The panel was generated by exporting the final checkpoint embeddings on all 12,285 held-out test patches, computing nearest-neighbor rationale and local/global retrieval, and then selecting visually diverse examples with a mix of successes and confounds. This is important because the aggregate metrics alone hide the shape of the problem: image-to-text recall@10 is only \(0.3787\), while text-to-image recall@10 is \(0.9161\). That asymmetry is expected when 12,285 patches share only 244 unique observation-level rationales. A text query often has many plausible image positives, but a single image can be close to several generic Olympus Mons rationales.

The successful rows show that the embedding is not random: crater, lava-flow, caldera, and basal-scarp rationales often retrieve visually plausible HiRISE texture and the paired global crop is frequently ranked first. The failure rows are equally useful. They show that dominant regional words such as ``Olympus Mons,'' ``lava,'' ``crater,'' and ``channel'' can pull together semantically adjacent but geologically different samples. The takeaway is therefore bounded: Stage B has learned a useful coarse Mars search space, but not yet a patch-specific geological captioner. The next research step is to reduce this text bottleneck with stronger patch-level descriptions, harder negative mining among visually similar Olympus terrains, and overlap-matched local/global controls.

\section{Dataset and Calibration Details}

The dataset layer is a major project contribution because HiRISE is not a ready-to-train machine-learning dataset. Each JP2 product has per-observation map projection metadata and must be interpreted in a Mars-specific coordinate system. We use a Mars IAU 2000 geographic CRS derived from the HiRISE RDR metadata. The global Mars ellipsoid uses equatorial radius \(a = 3,396,190\) m and polar radius \(b = 3,376,200\) m~\cite{DSMAP}. Each observation records a local radius at the center latitude:
\[
R = \frac{a b}{\sqrt{\left(b \cos(\mathrm{LatP})\right)^2 + \left(a \sin(\mathrm{LatP})\right)^2}}.
\]

HiRISE imagery is provided as raw digital numbers rather than physical reflectance. Each product supplies scaling and offset metadata, allowing conversion to radiance factor:
\[
    \mathrm{I/F} = \mathrm{DN} \times \mathrm{SCALING\_FACTOR} + \mathrm{OFFSET}.
\]
This calibration is applied before patch extraction so training sees physically meaningful image values rather than product-specific integer ranges.

\begin{table}[!htbp]
\centering
\caption{\textbf{HiRISE spectral bands used by the MarsRecon data pipeline.} The RED channel provides the highest-resolution structural texture signal, while blue-green and near-infrared channels provide additional spectral context where color products are available~\cite{mcewen2010hirise,Sutton2022RevealingModels}.}
\label{tab:sensor_characteristics_appendix}
\small
\begin{tabular}{@{}lcp{8.1cm}@{}}
\toprule
\textbf{Band} & \textbf{Wavelength range} & \textbf{Primary use} \\
\midrule
Blue--Green & 400--600 nm & Frost, ice, atmospheric scattering, and color context \\
RED & 550--850 nm & High-resolution geomorphology, structural texture, and topographic interpretation \\
Near-Infrared & 800--1000 nm & Mineralogical differentiation and spectral contrast where available \\
\bottomrule
\end{tabular}
\end{table}

\begin{table}[!htbp]
\centering
\caption{\textbf{Radiometrically calibrated dataset statistics.} Statistics are computed over the Olympus patch corpus after DN-to-I/F conversion. P2/P98 are the 2nd and 98th percentiles, CV is the coefficient of variation, and DRCR is \((P_{98}-P_{2})/\mu\).}
\label{tab:dataset_stats_extended_appendix}
\scriptsize
\begin{tabular}{lrrrrrrrr}
\toprule
\textbf{Channel} & \textbf{\(N_{\text{pixels}}\)} & \textbf{Mean} & \textbf{Std} & \textbf{Min} & \textbf{Max} & \textbf{P2} & \textbf{P98} & \textbf{CV / DRCR} \\
\midrule
Near-infrared & \(4.5949{\times}10^{10}\) & 0.1965 & 0.0569 & 0.0097 & 0.7700 & 0.0928 & 0.3125 & 0.290 / 1.118 \\
RED & \(4.9983{\times}10^{10}\) & 0.1668 & 0.0391 & 0.0097 & 0.3600 & 0.0879 & 0.2471 & 0.235 / 0.954 \\
Blue-green & \(4.6060{\times}10^{10}\) & 0.0723 & 0.0211 & 0.0097 & 0.3184 & 0.0389 & 0.1270 & 0.292 / 1.218 \\
\bottomrule
\end{tabular}
\end{table}

\begin{figure}[!htbp]
     \centering
     \begin{subfigure}[b]{0.32\textwidth}
         \centering
         \includegraphics[width=\textwidth]{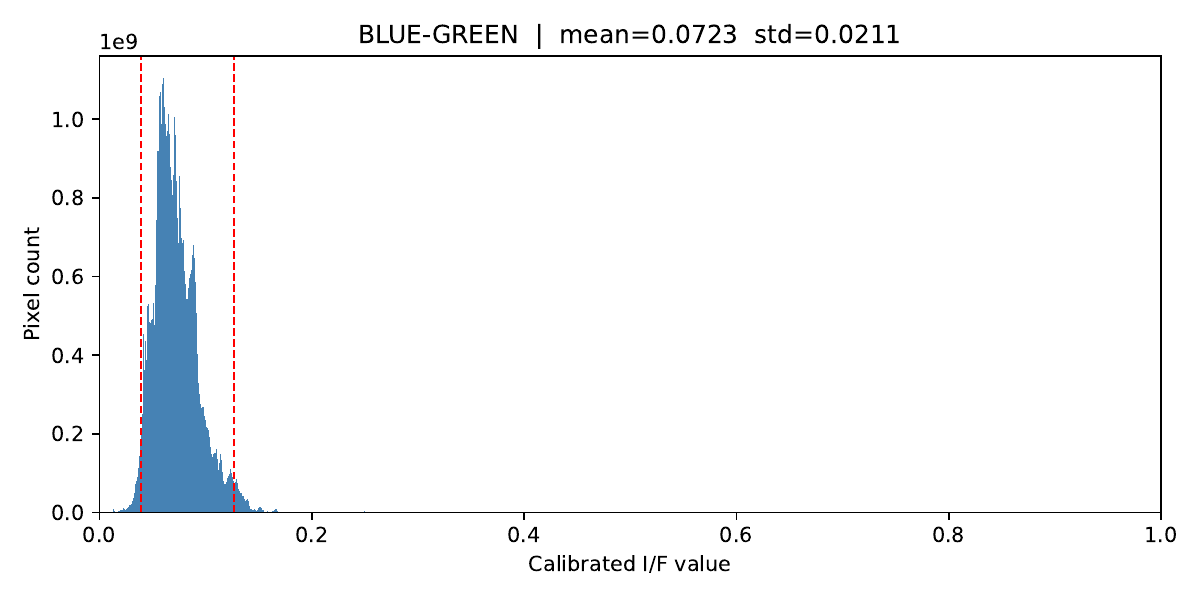}
         \caption{Blue-green}
     \end{subfigure}
     \hfill
     \begin{subfigure}[b]{0.32\textwidth}
         \centering
         \includegraphics[width=\textwidth]{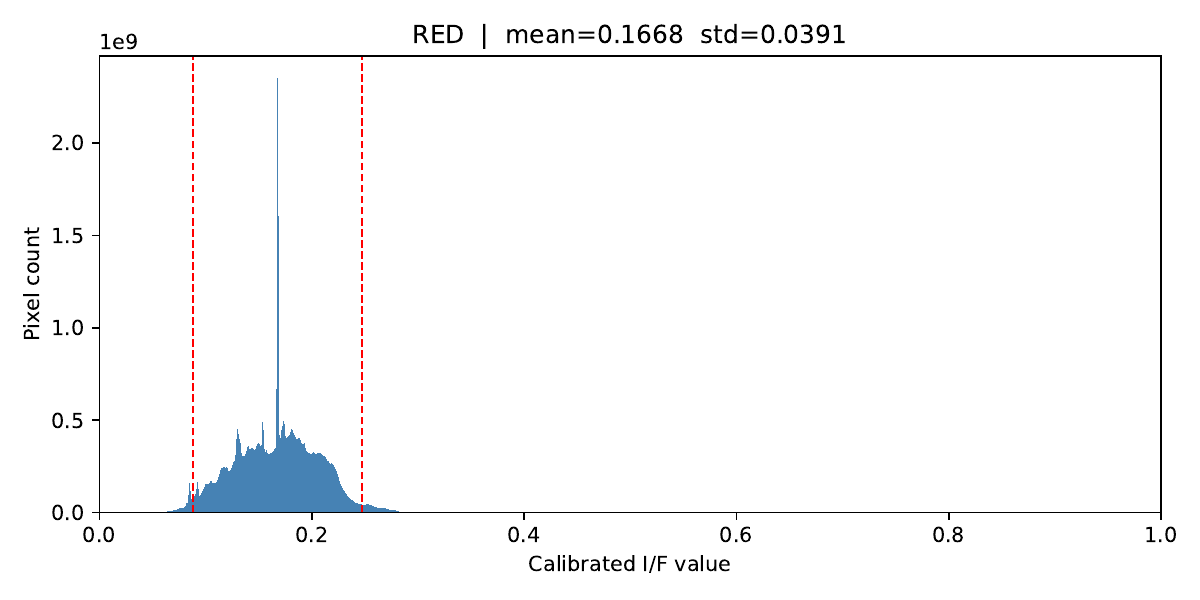}
         \caption{RED}
     \end{subfigure}
     \hfill
     \begin{subfigure}[b]{0.32\textwidth}
         \centering
         \includegraphics[width=\textwidth]{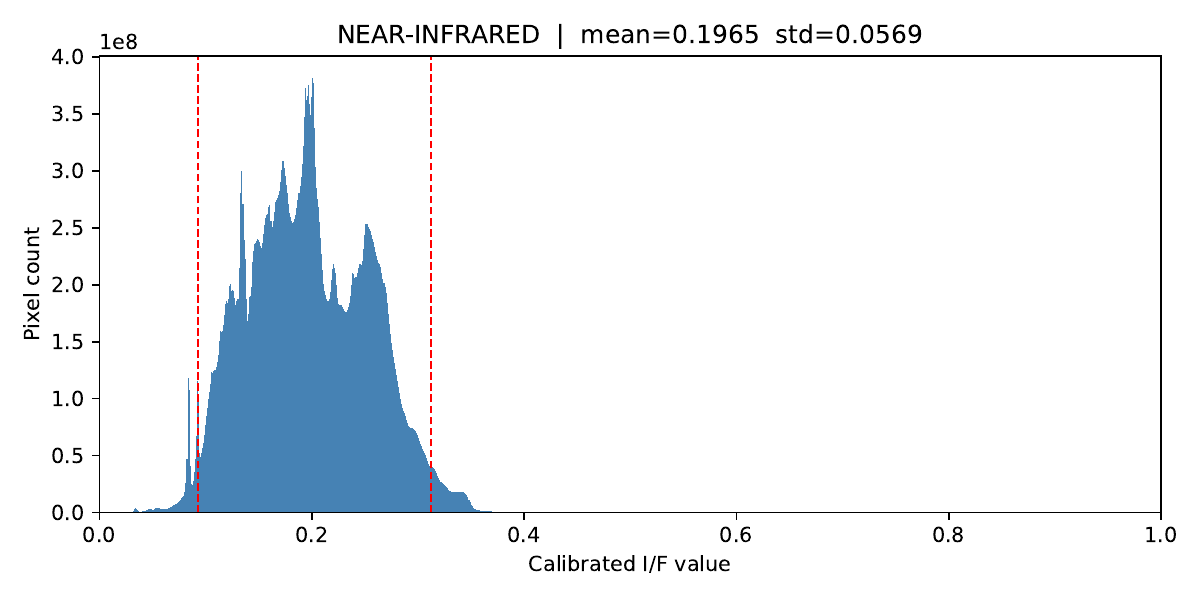}
         \caption{Near-infrared}
     \end{subfigure}
     \caption{\textbf{Radiometrically calibrated pixel distributions.} The histograms summarize calibrated I/F values for the Olympus patch corpus and show why dataset-level normalization and percentile-aware visualization are useful for training and qualitative inspection.}
     \label{fig:pixel_histograms_appendix}
\end{figure}

\begin{table}[!htbp]
\centering
\caption{\textbf{Representative PDS metadata fields used by the data pipeline.} These fields make each patch geospatially and physically interpretable rather than a detached image crop.}
\label{tab:pds_metadata_complete_appendix}
\scriptsize
\renewcommand{\arraystretch}{1.13}
\begin{tabular}{@{}lp{3.0cm}p{8.0cm}@{}}
\toprule
\textbf{Parameter} & \textbf{Example value} & \textbf{Role in MarsRecon} \\
\midrule
\textbf{OBSERVATION\_ID} & TRA\_000873\_2015 & Unique orbital capture identifier. \\
\textbf{PRODUCT\_ID} & TRA\_000873\_COLOR & Product-specific identifier, including color/RED distinction. \\
\textbf{OBSERVATION\_START\_TIME} & 2006-10-03T12:51:33.015 & Timestamp for temporal and mission-phase context. \\
\textbf{MAP\_PROJECTION\_TYPE} & EQUIRECTANGULAR & Per-product map projection used before transformation into the common Mars geographic CRS. \\
\textbf{MAP\_SCALE} & 0.25 m/pixel & Spatial resolution at the projected center latitude. \\
\textbf{CENTER\_LATITUDE / LONGITUDE} & 20.000\(^{\circ}\), 180.000\(^{\circ}\) & Planetocentric map center and coordinate context. \\
\textbf{INCIDENCE\_ANGLE} & 47.099670\(^{\circ}\) & Solar zenith angle; relevant for illumination and surface scattering. \\
\textbf{EMISSION\_ANGLE} & 0.282987\(^{\circ}\) & Instrument line-of-sight relative to local surface normal. \\
\textbf{PHASE\_ANGLE} & 47.047614\(^{\circ}\) & Sun-target-observer angle, useful for BRDF/viewing-condition analysis. \\
\textbf{LOCAL\_TIME} & 15.40208 & Local solar time of the observation. \\
\textbf{SOLAR\_LONGITUDE} & 115.346344\(^{\circ}\) & Areocentric solar longitude \(L_s\), encoding seasonal context. \\
\textbf{SCALING\_FACTOR / OFFSET} & \(2.1639{\times}10^{-4}\) / 0.0461 & Linear coefficients for DN-to-I/F radiometric calibration. \\
\bottomrule
\end{tabular}
\end{table}

\section{Olympus Subset and Patch Validation}

\begin{figure}[!htbp]
    \centering
    \includegraphics[width=\linewidth]{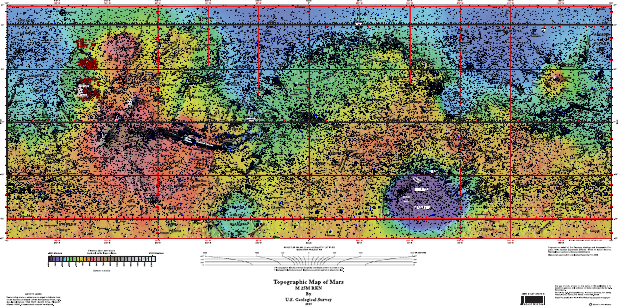}
    \caption{\textbf{HiRISE coverage and Olympus subset.} Blue points show available HiRISE sample locations in the plotted range, while red points show the project subset selected around Olympus Mons. The base topographic map is from USGS Mars topography products~\cite{Topographic2782}.}
    \label{fig:toptographic_map_appendix}
\end{figure}

The Olympus Mons subset was chosen to make the project computationally tractable while preserving geological diversity. Rather than downloading the full HiRISE archive, the project filters relevant products, builds valid patch records, and splits them into train/validation/test partitions. Figure~\ref{fig:toptographic_map_appendix} shows the global context of the selected observations.

Patch validation is necessary because HiRISE strips are long, angled, and often surrounded by nodata regions inside their metadata bounding boxes. The pipeline therefore estimates valid data support and rejects patches that are dominated by fill pixels.

\begin{figure}[!htbp]
    \centering
    \includegraphics[width=\linewidth]{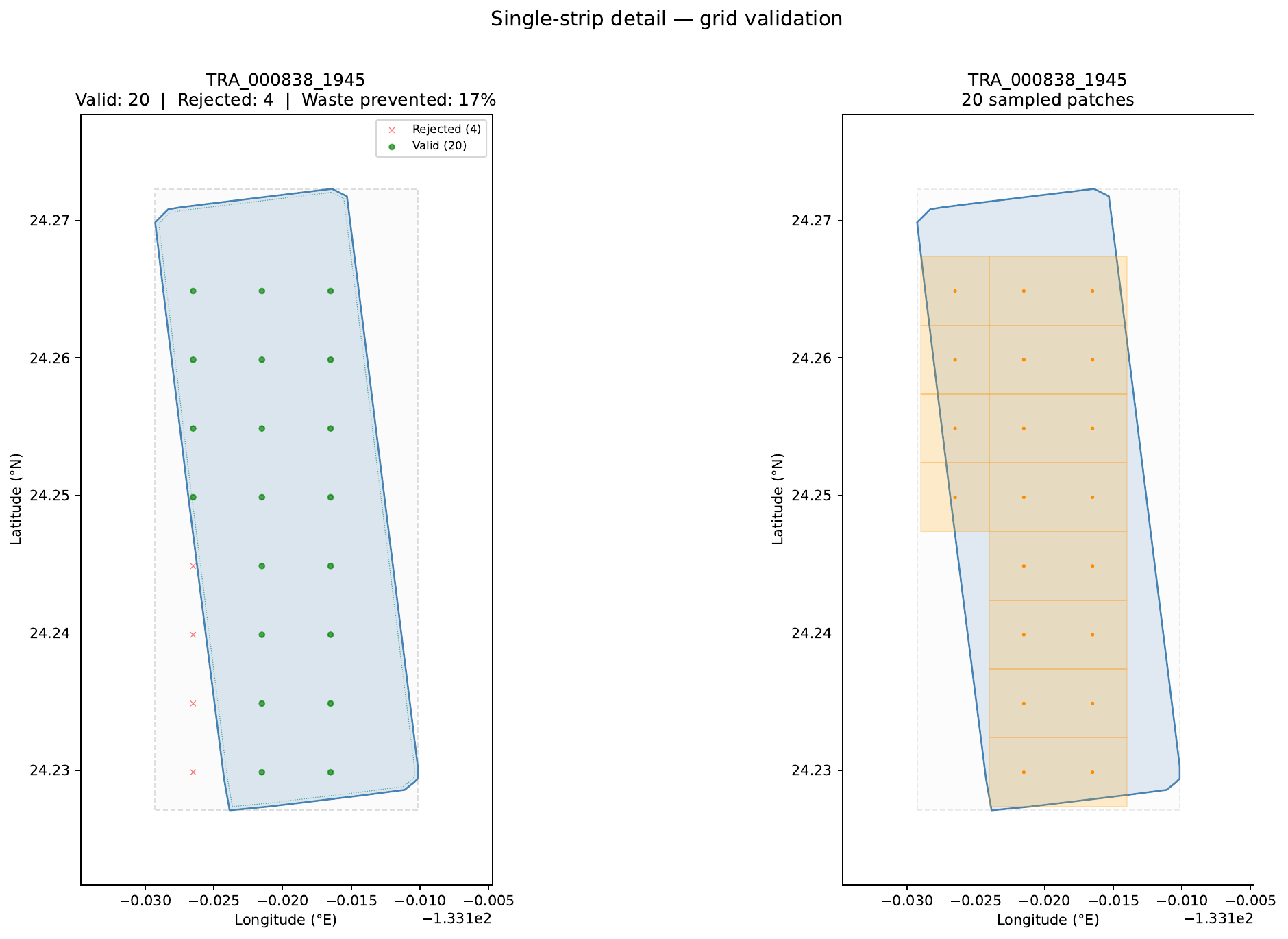}
    \caption{\textbf{Valid-support estimation for a HiRISE strip.} The gray rectangle is the metadata bounding box, while the valid image footprint is narrower and angled. Valid-mask-aware sampling prevents the model from learning artificial nodata patterns.}
    \label{fig:strip_validation_appendix}
\end{figure}

\begin{figure}[!htbp]
    \centering
    \includegraphics[width=\linewidth]{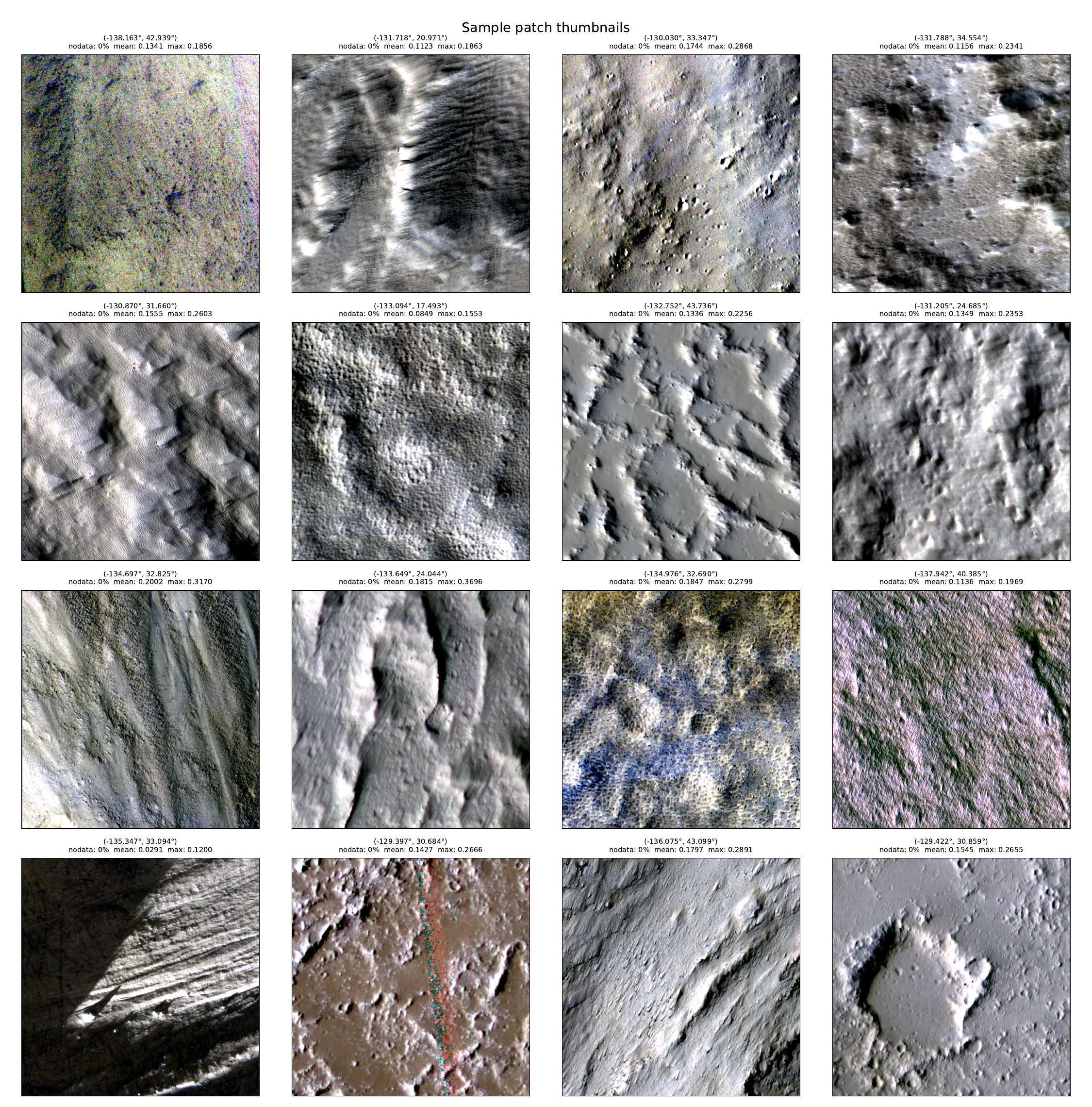}
    \caption{\textbf{Example Olympus patches.} Random examples from the selected subset illustrate the range of surface textures, illumination conditions, and strip geometries seen during training.}
    \label{fig:thumbnail_examples_appendix}
\end{figure}

\section{Rationale Text Analysis}

The HiRISE index contains a \texttt{RATIONALE\_DESC} field explaining why each observation was collected. The following analysis parses the official PDS fixed-width index, which enforces a 75 character limit, deduplicates repeated RED/COLOR products by observation ID, computes observation centers from latitude/longitude bounds, tokenizes the rationale text, removes non-scientific stopwords, preserves hyphenated geological terms, and assigns science themes through keyword sets tuned to HiRISE vocabulary. This analysis motivates Stage B: the text is scientifically meaningful, but it is coarse and repetitive enough that false-negative-aware contrastive learning is necessary.

\begin{figure}[!htbp]
    \centering
    \begin{subfigure}[b]{0.49\linewidth}
        \centering
        \includegraphics[width=\linewidth]{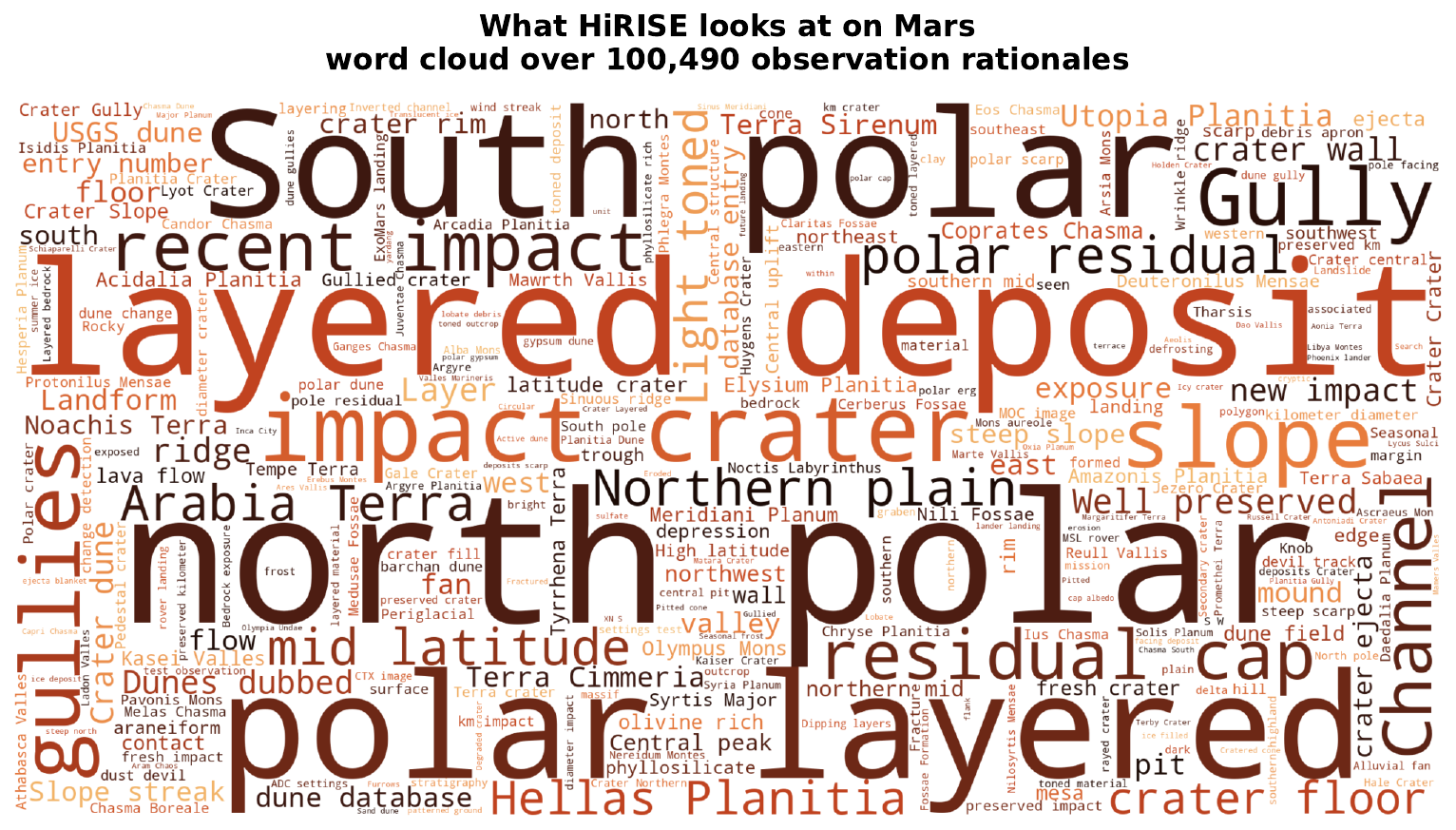}
        \caption{Full word cloud}
    \end{subfigure}
    \hfill
    \begin{subfigure}[b]{0.39\linewidth}
        \centering
        \includegraphics[width=\linewidth]{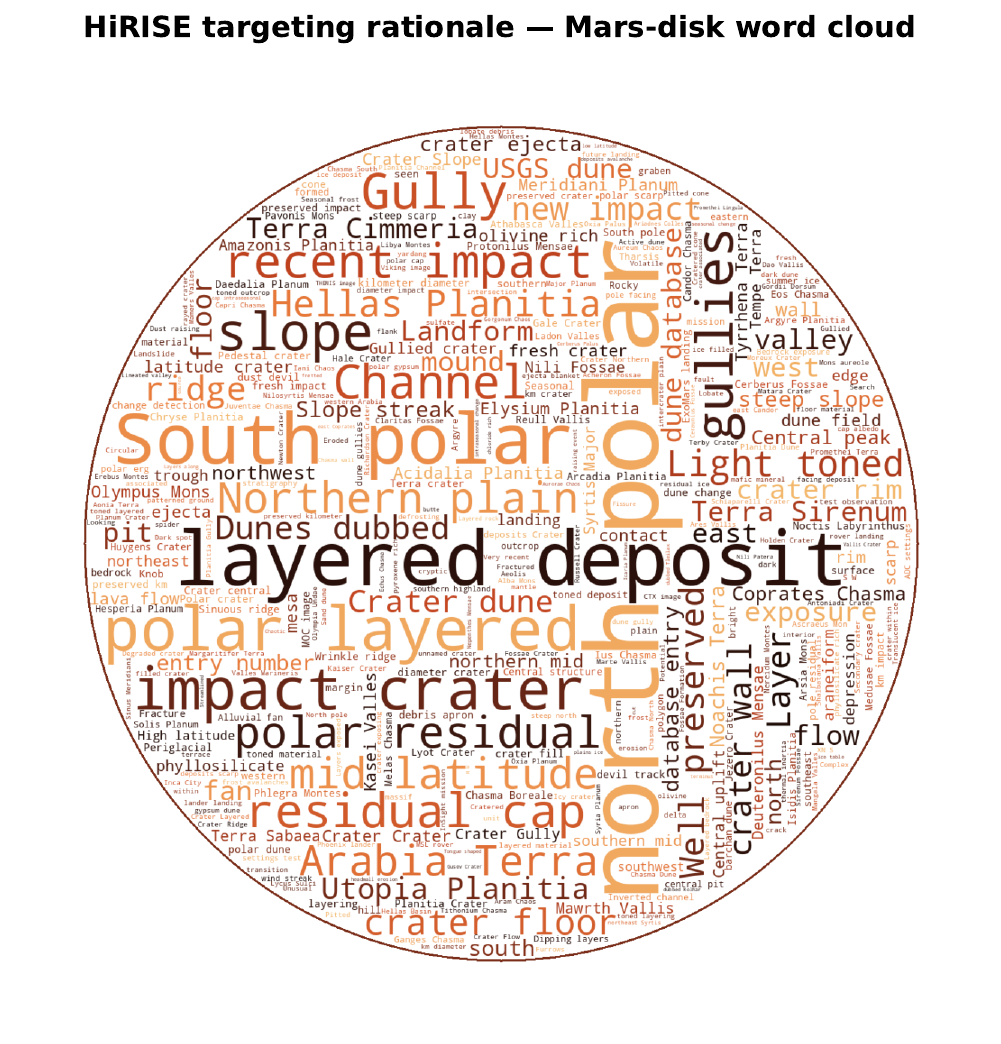}
        \caption{Mars-disk visual variant}
    \end{subfigure}
    \caption{\textbf{HiRISE targeting vocabulary.} The word clouds summarize the most prominent terms in observation rationales after stopword filtering. They show that HiRISE text is strongly geologic rather than generic metadata, with repeated concepts such as craters, dunes, layers, polar processes, slopes, deposits, and channels.}
    \label{fig:rationale_wordclouds_appendix}
\end{figure}

\begin{figure}[!htbp]
    \centering
    \includegraphics[width=\linewidth]{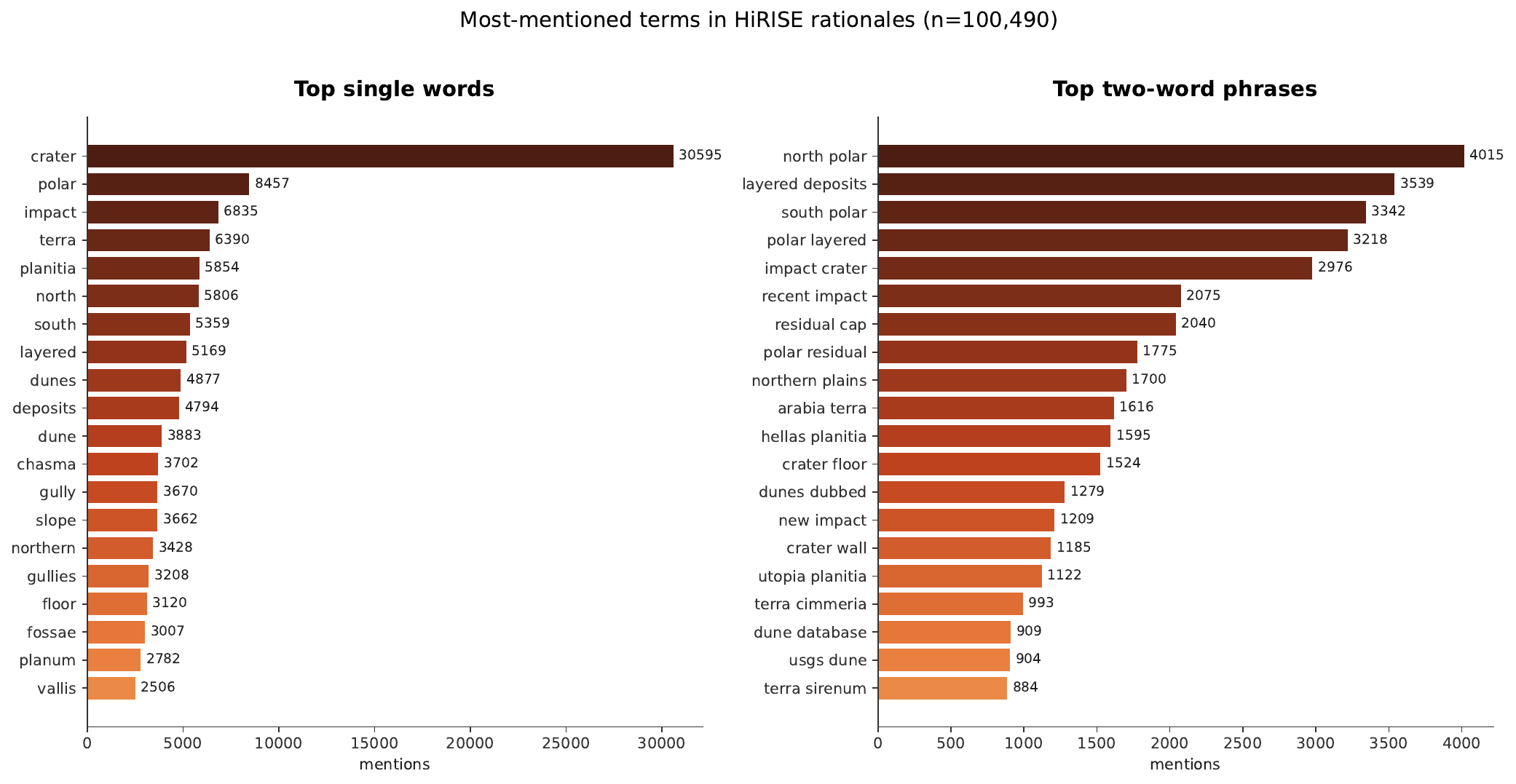}
    \caption{\textbf{Top rationale words and phrases.} The script counts filtered unigrams and adjacent token bigrams, making the text distribution auditable rather than relying only on the word-cloud visualization.}
    \label{fig:rationale_top_words_appendix}
\end{figure}

\begin{figure}[!htbp]
    \centering
    \begin{subfigure}[b]{0.49\linewidth}
        \centering
        \includegraphics[width=\linewidth]{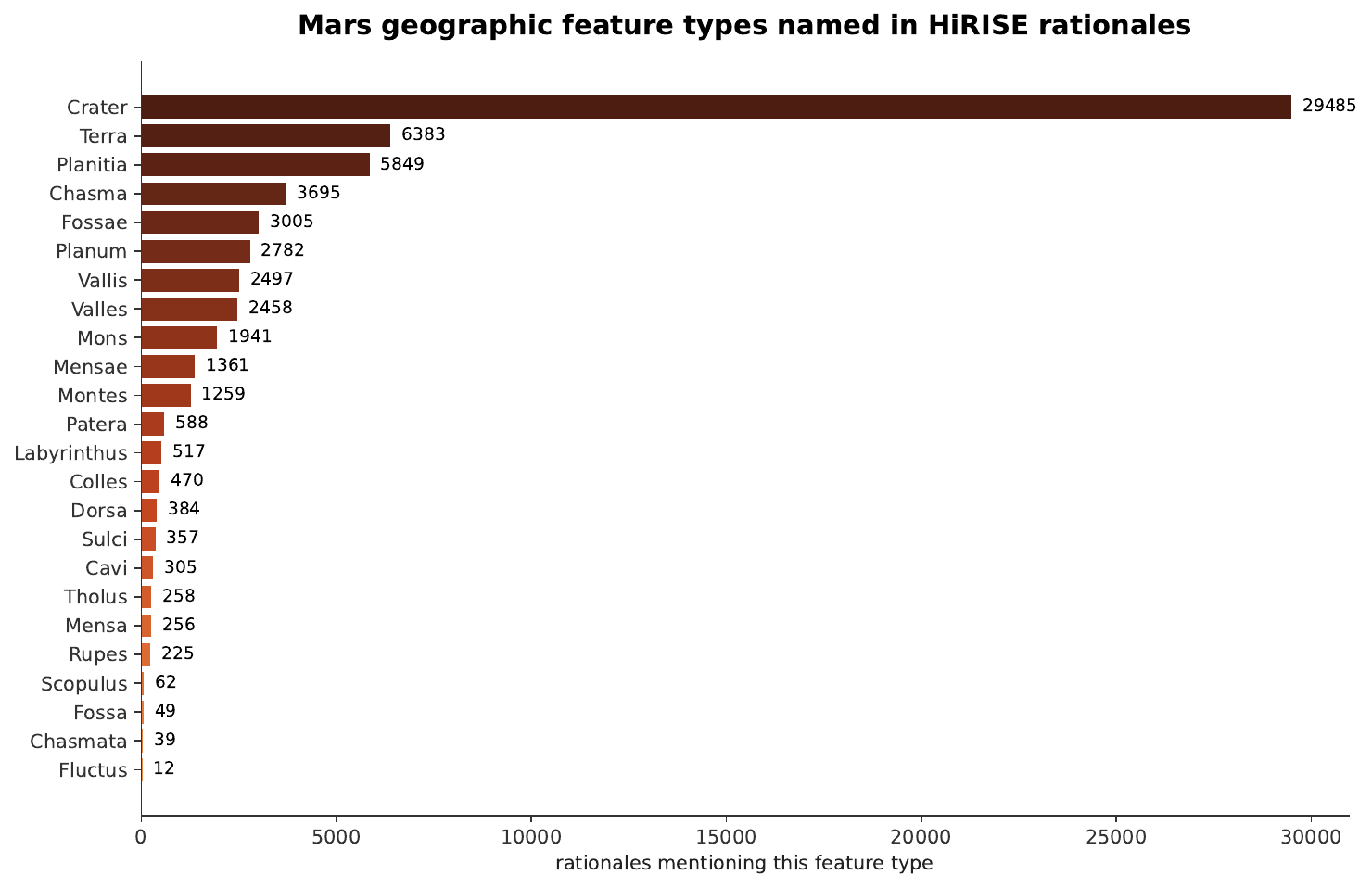}
        \caption{Named geographic feature types}
    \end{subfigure}
    \hfill
    \begin{subfigure}[b]{0.49\linewidth}
        \centering
        \includegraphics[width=\linewidth]{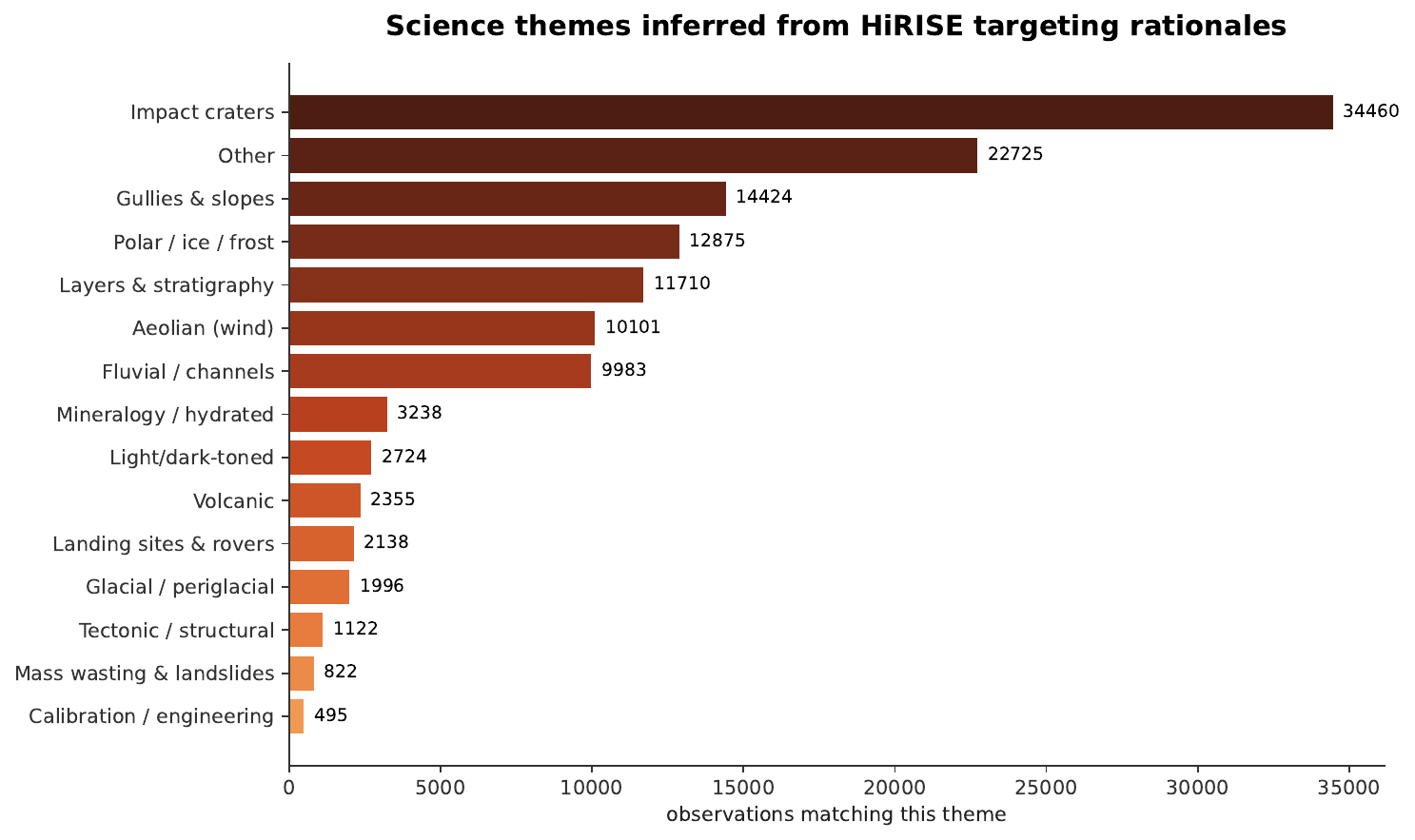}
        \caption{Keyword-derived science themes}
    \end{subfigure}
    \caption{\textbf{From words to geological categories.} Geographic-feature counts track named Mars feature types such as craters, chasmata, valles, fossae, and montes. Theme counts group rationales into science categories including impact craters, gullies/slopes, aeolian processes, polar/ice/frost, layers/stratigraphy, mineralogy, volcanic processes, tectonics, and landing-site/engineering observations.}
    \label{fig:rationale_theme_counts_appendix}
\end{figure}

\begin{figure}[!htbp]
    \centering
    \includegraphics[width=\linewidth]{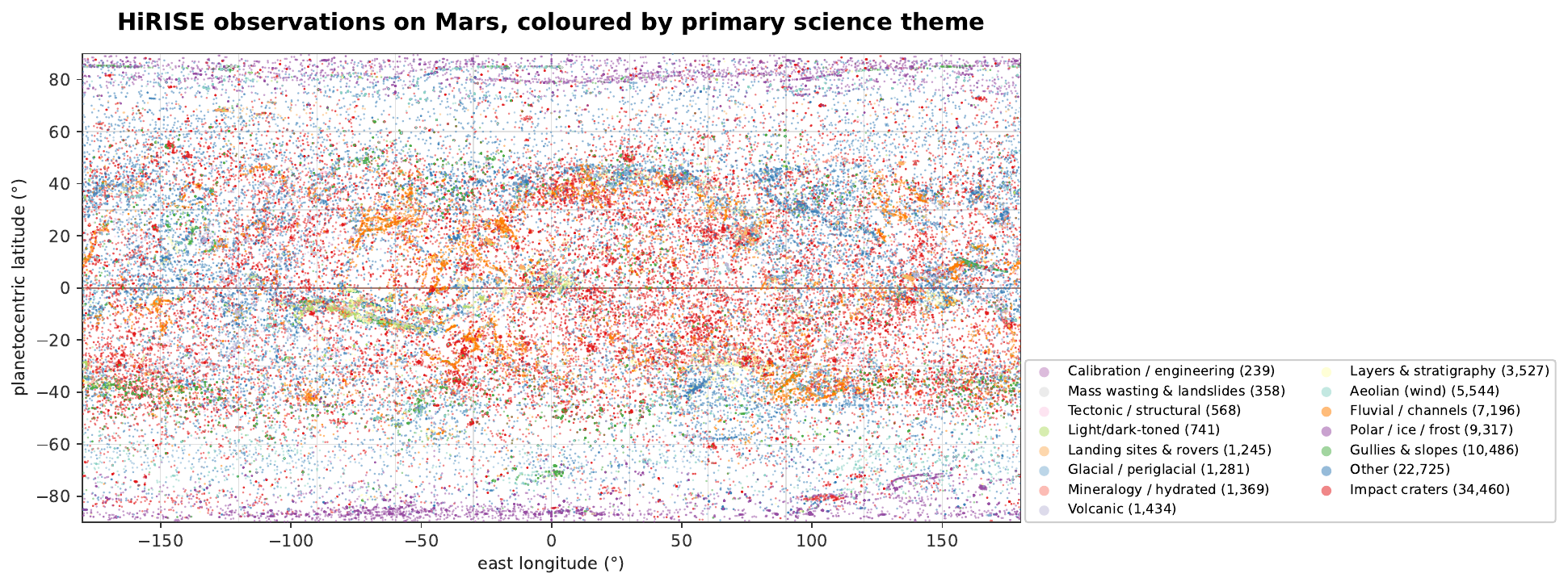}
    \caption{\textbf{Observation map colored by primary science theme.} Each observation is assigned a primary theme from its matched keyword list and plotted by longitude/latitude. This connects the Stage B text signal back to physical Mars geography.}
    \label{fig:rationale_themed_map_appendix}
\end{figure}

\begin{figure}[!htbp]
    \centering
    \includegraphics[width=\linewidth]{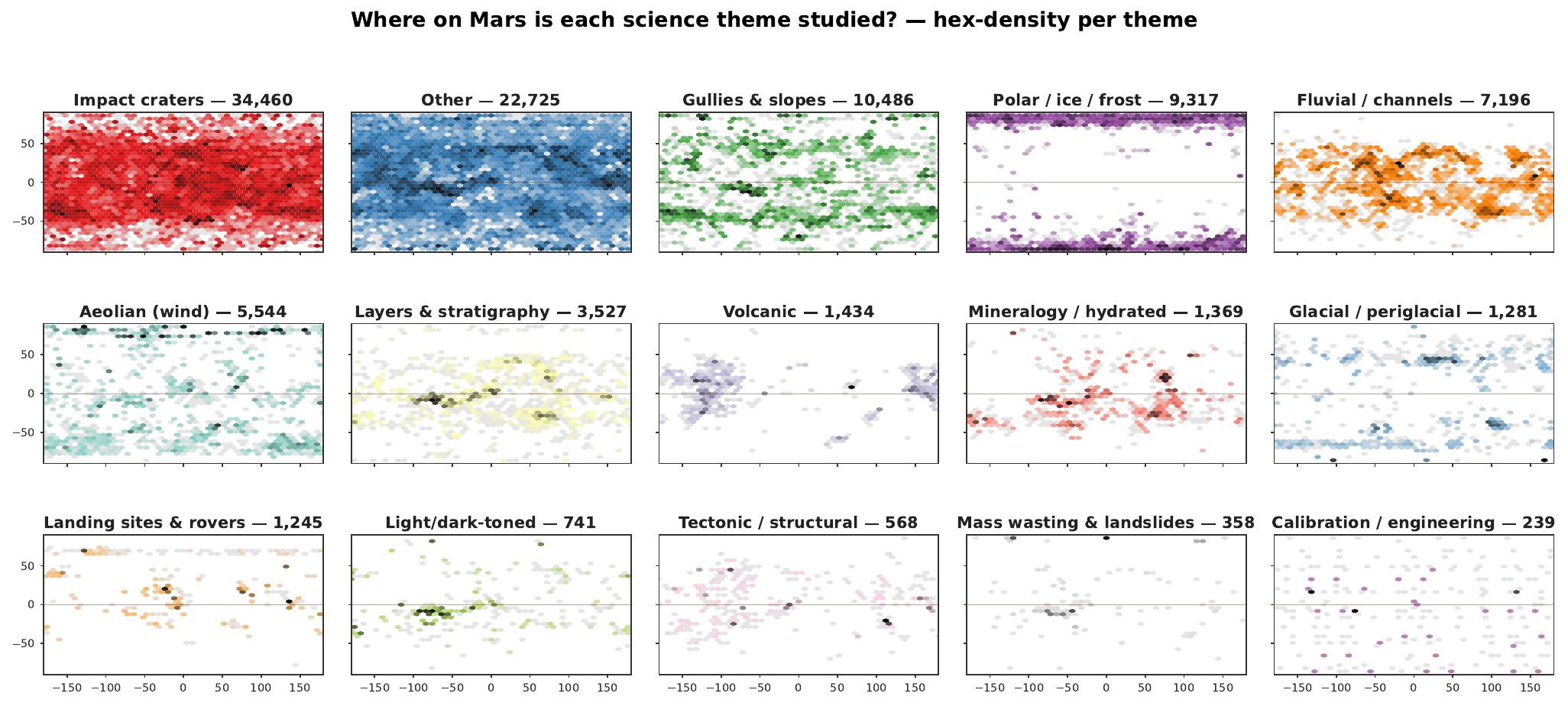}
    \caption{\textbf{Theme-specific Mars maps.} Small multiples reduce overplotting by drawing one hex-density map per science theme, revealing where different rationale classes cluster spatially.}
    \label{fig:rationale_smallmultiples_appendix}
\end{figure}

\begin{figure}[!htbp]
    \centering
    \includegraphics[width=\linewidth]{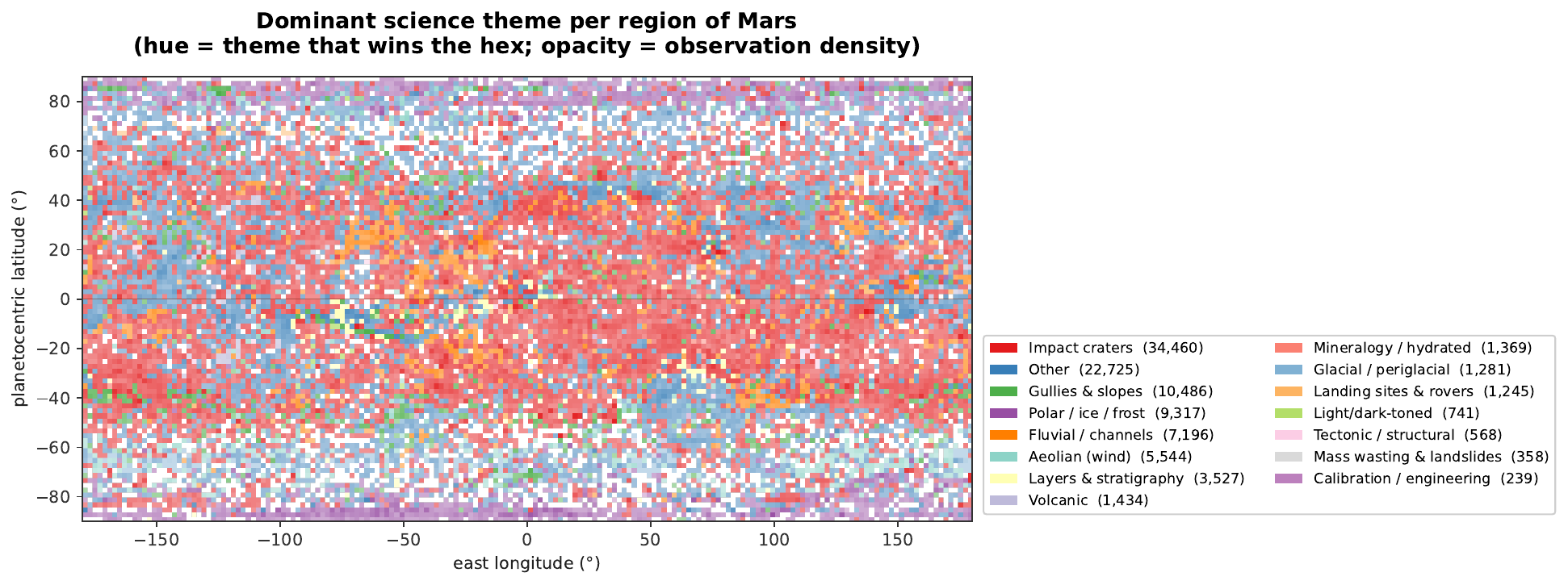}
    \caption{\textbf{Dominant theme per Mars region.} Each spatial bin is colored by the locally dominant primary theme, with opacity scaled by log observation count. This preserves a single-map view while reducing the overplotting problem of point-wise thematic maps.}
    \label{fig:rationale_dominant_map_appendix}
\end{figure}

\begin{figure}[!htbp]
    \centering
    \begin{subfigure}[b]{0.46\linewidth}
        \centering
        \includegraphics[width=\linewidth]{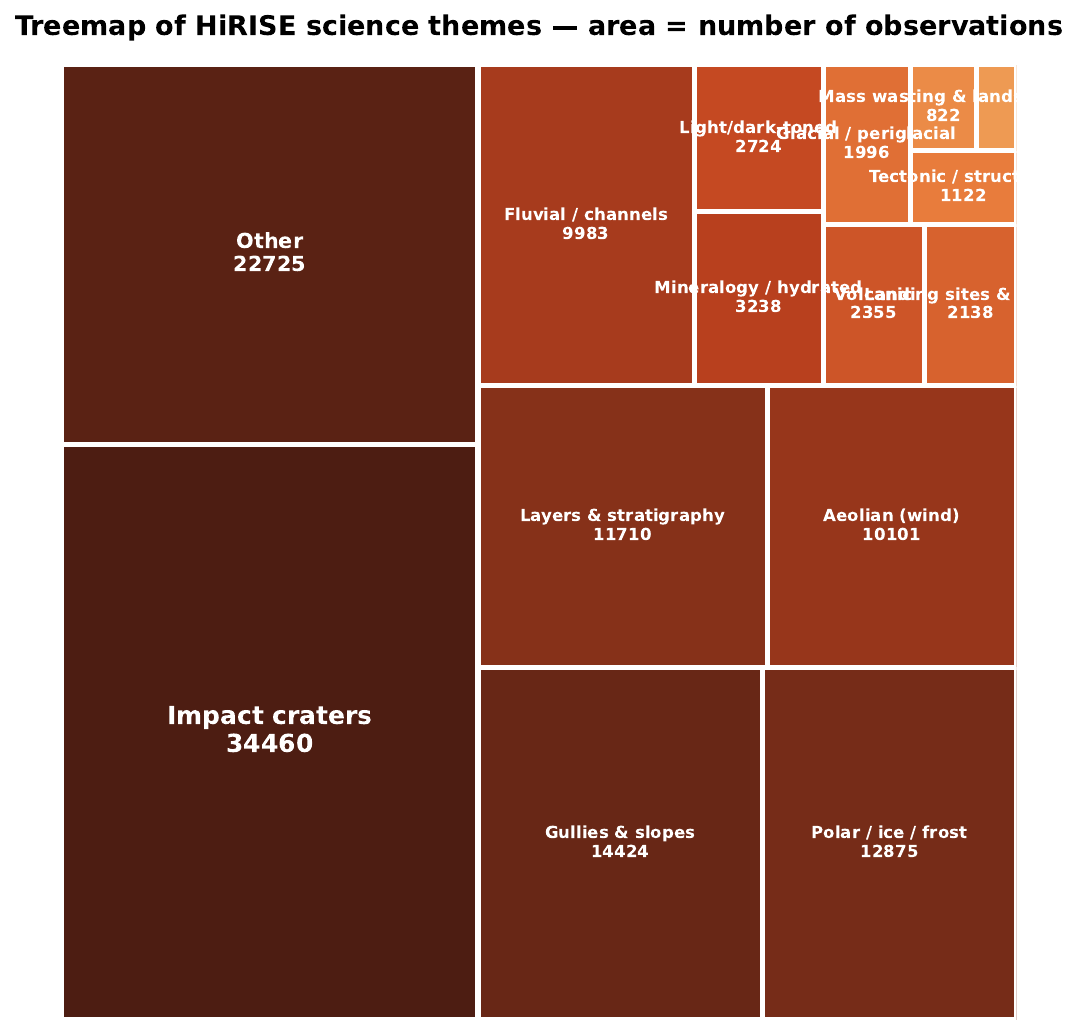}
        \caption{Theme proportions}
    \end{subfigure}
    \hfill
    \begin{subfigure}[b]{0.50\linewidth}
        \centering
        \includegraphics[width=\linewidth]{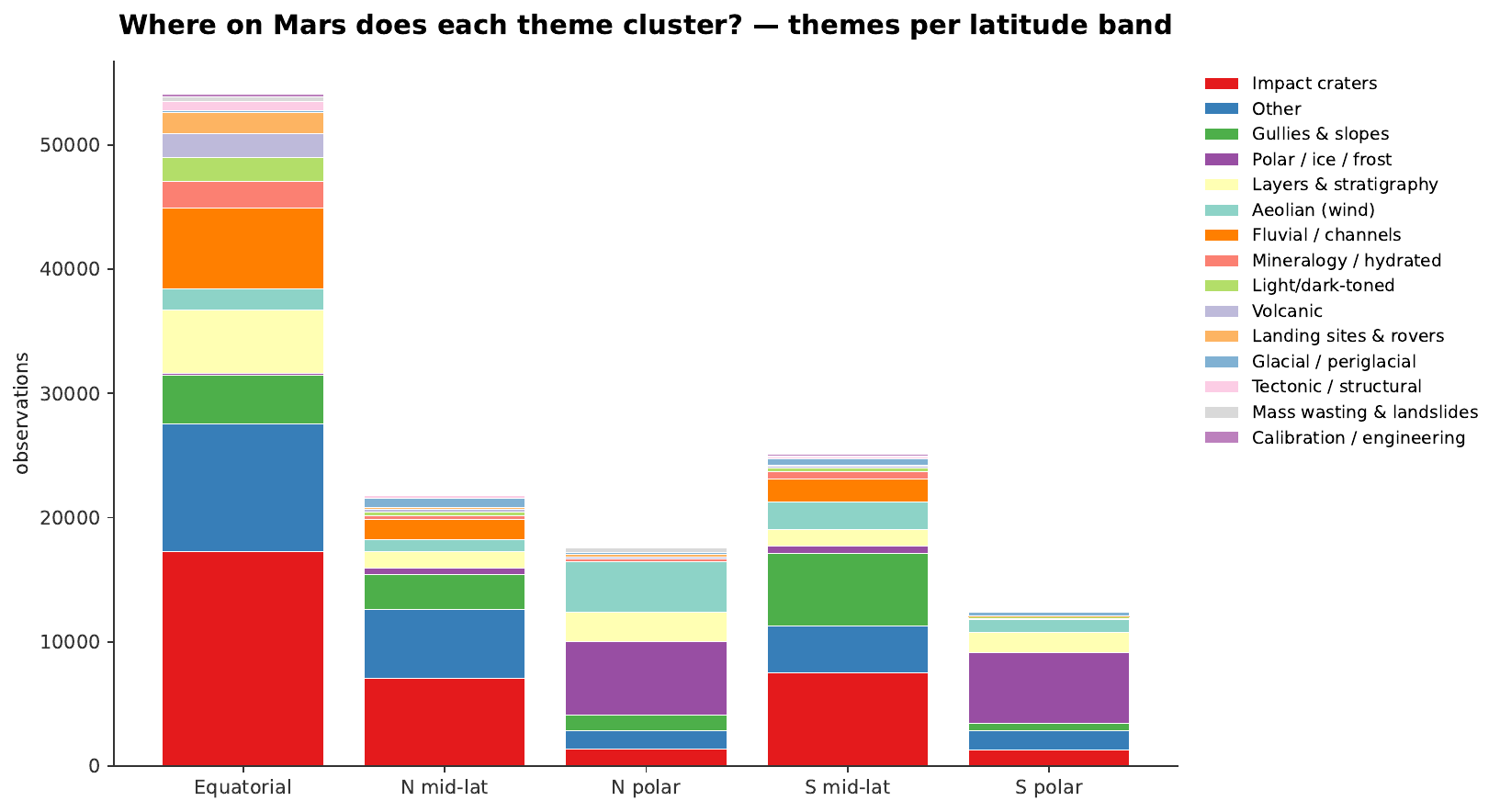}
        \caption{Themes by latitude band}
    \end{subfigure}
    \caption{\textbf{Theme imbalance and latitude dependence.} The treemap shows the global class imbalance in rationale-derived science themes, while the stacked latitude-band plot shows that theme frequencies vary by equatorial, mid-latitude, and polar regions. This helps explain why Stage B should combine text with location rather than treating rationales as globally exchangeable labels.}
    \label{fig:rationale_theme_distribution_appendix}
\end{figure}

\section{Sampling and Preprocessing}

We use boundary-aware sampling because simple axis-aligned grids waste many candidate patches along angled HiRISE strips. Given a convex valid footprint \(P\), an axis-aligned square patch \(S(c)\) of side length \(L\), and a required footprint-overlap fraction \(p\), valid patch centers are defined as
\[
R = \{ c \in \mathbb{R}^2 \mid \mathrm{Area}(P \cap S(c)) \ge p L^2 \}.
\]
Within each horizontal strip of the valid center region, patch centers are spaced by stride \(\sigma=L(1-\delta)\), where \(\delta\) is the allowed patch-overlap fraction. For a continuous 1D strip segment of width \(W\), the maximum number of centers is
\[
N_{\max} = \left\lfloor \frac{W}{\sigma} \right\rfloor + 1.
\]
The final Stage A/B experiments use patch records produced by this broader valid-support infrastructure.

\begin{figure}[!htbp]
    \centering
    \includegraphics[width=\linewidth]{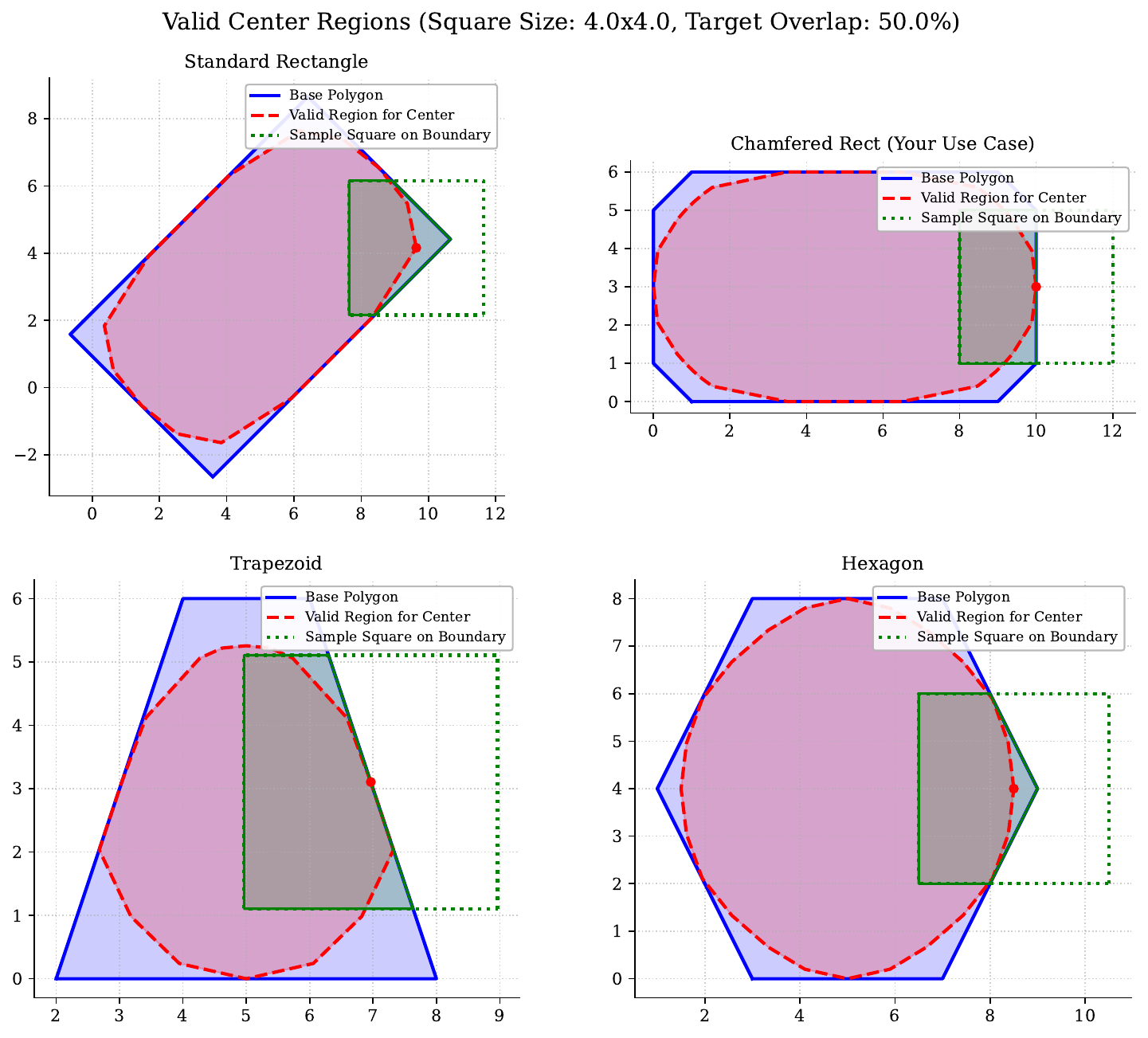}
    \caption{\textbf{Valid-center region from overlap constraints.} The green polygon shows where patch centers can be placed while maintaining the minimum required overlap with the valid strip footprint. This converts a nodata-prone bounding-box problem into a constrained sampling problem.}
    \label{fig:min_overlap_polygon_appendix}
\end{figure}

\begin{figure}[!htbp]
    \centering
    \begin{subfigure}[b]{0.49\linewidth}
        \centering
        \includegraphics[width=\linewidth]{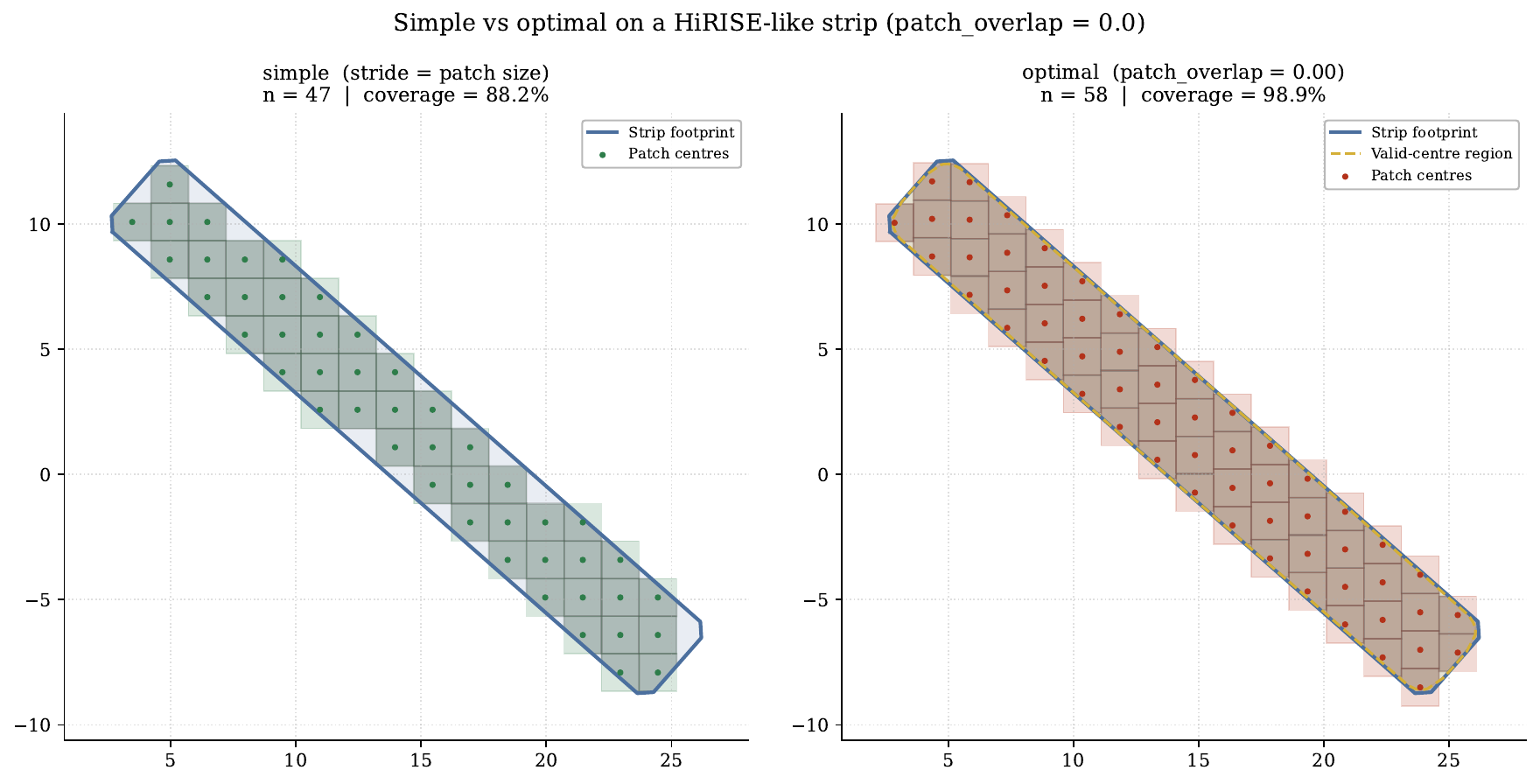}
        \caption{0\% patch-overlap stride}
    \end{subfigure}
    \hfill
    \begin{subfigure}[b]{0.49\linewidth}
        \centering
        \includegraphics[width=\linewidth]{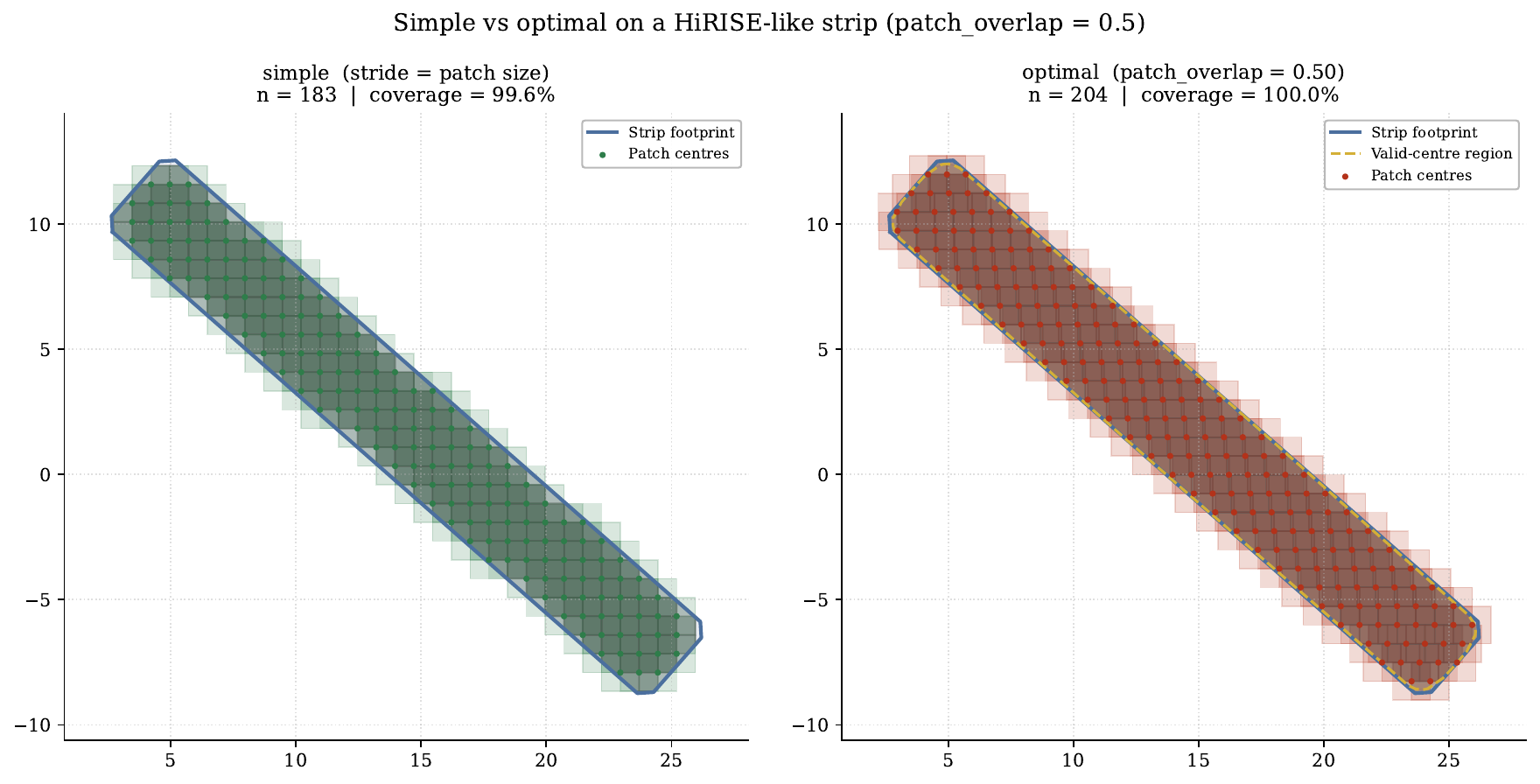}
        \caption{50\% patch-overlap stride}
    \end{subfigure}
    \caption{\textbf{Simple grid versus independent strip packing.} The strip-aware method adapts each row to the valid center region, improving valid patch yield over rigid column-locked grids for angled orbital strips.}
    \label{fig:hero_strip_appendix}
\end{figure}

\begin{figure}[!htbp]
    \centering
    \includegraphics[width=\linewidth]{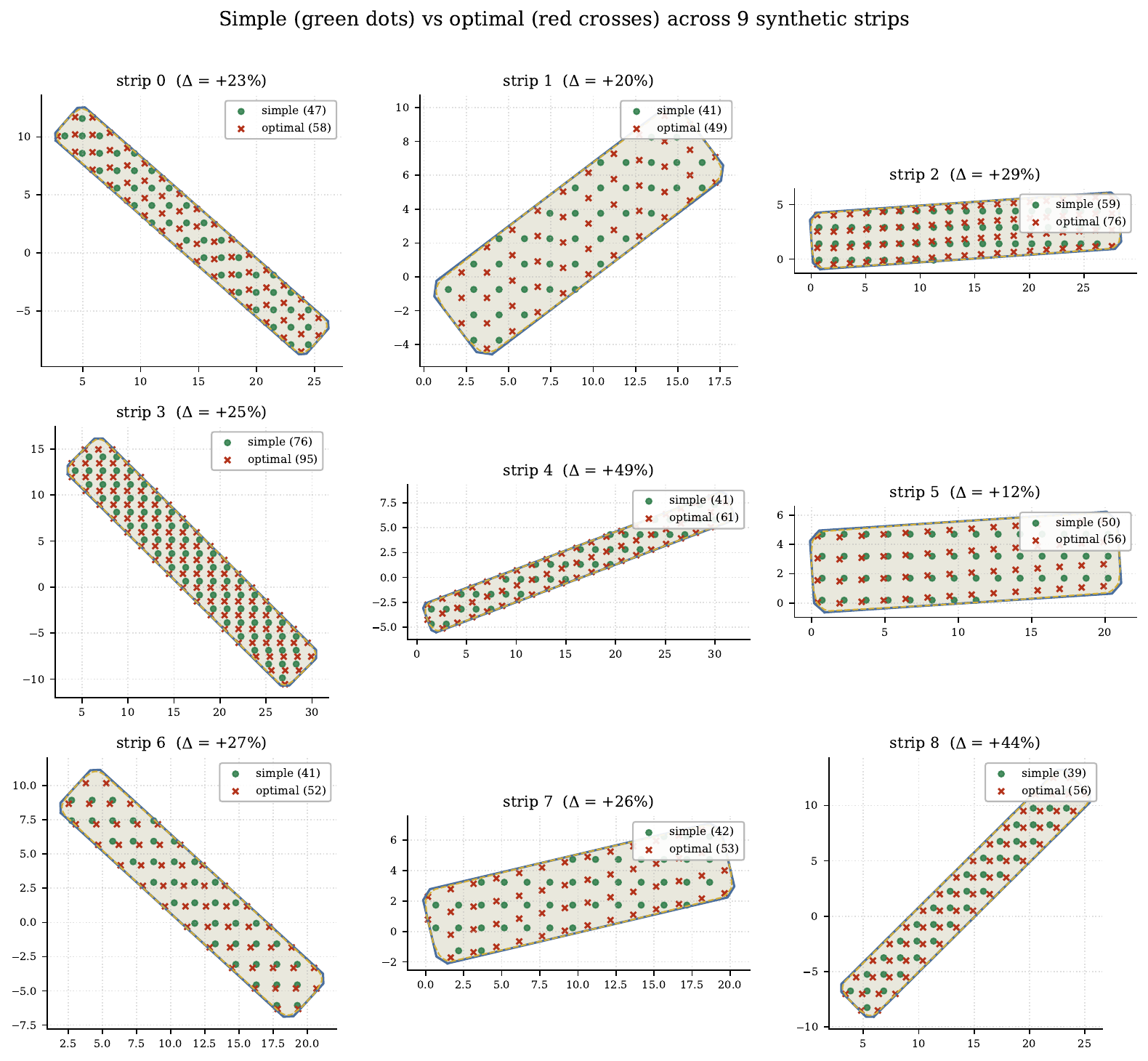}
    \caption{\textbf{Sampler robustness across strip geometries.} Across synthetic angled strip footprints, the boundary-aware method consistently finds more valid patch centers than a simple grid.}
    \label{fig:gallery_strips_appendix}
\end{figure}

\begin{figure}[!htbp]
    \centering
    \includegraphics[width=\linewidth]{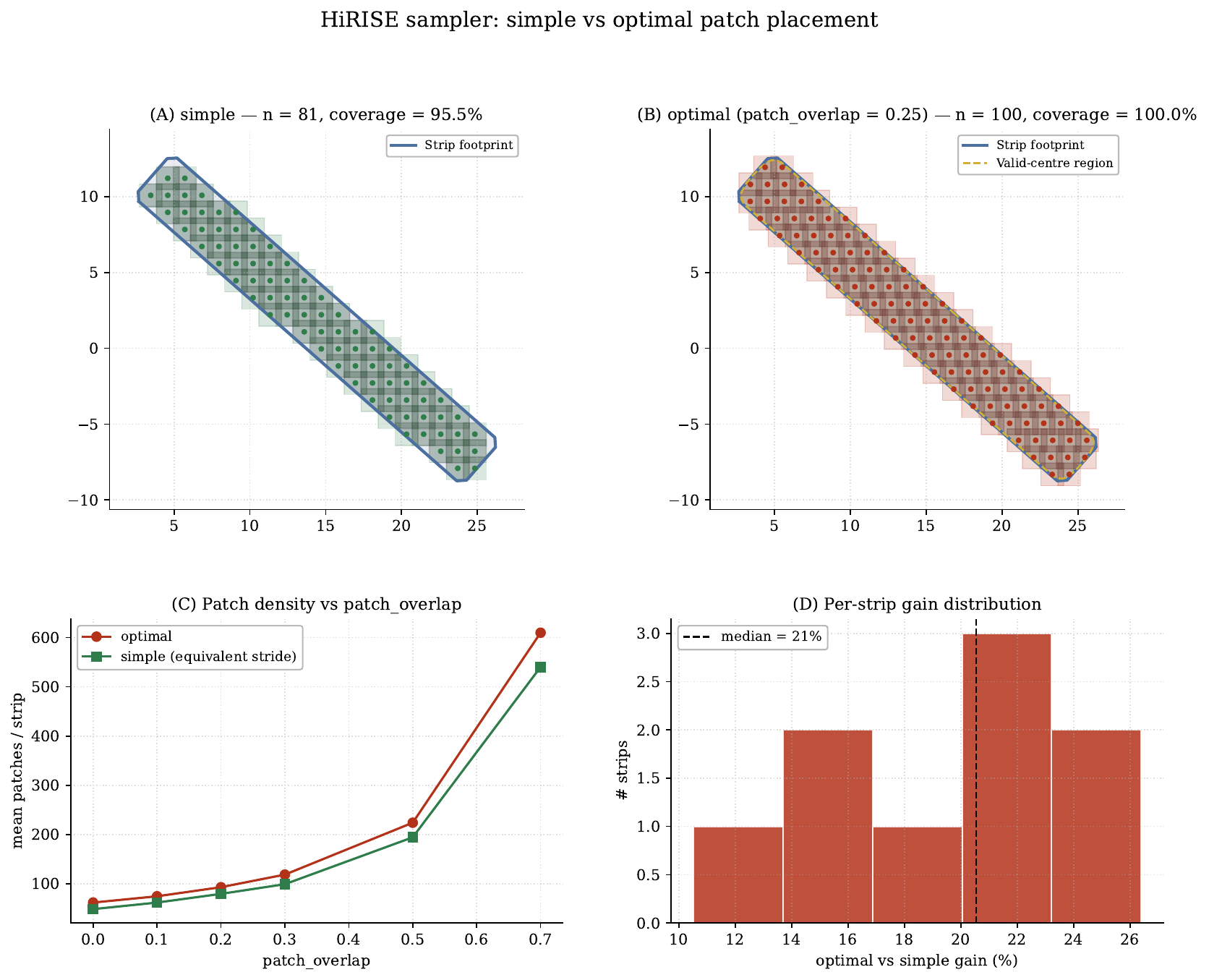}
    \caption{\textbf{Boundary-aware patch sampling.} The sampler increases valid patch density by adapting candidate centers to the valid strip footprint rather than relying on a rigid grid over the metadata bounding box.}
    \label{fig:publication_figure_appendix}
\end{figure}

\begin{figure}[!htbp]
    \centering
    \begin{subfigure}[b]{0.49\linewidth}
        \centering
        \includegraphics[width=\linewidth]{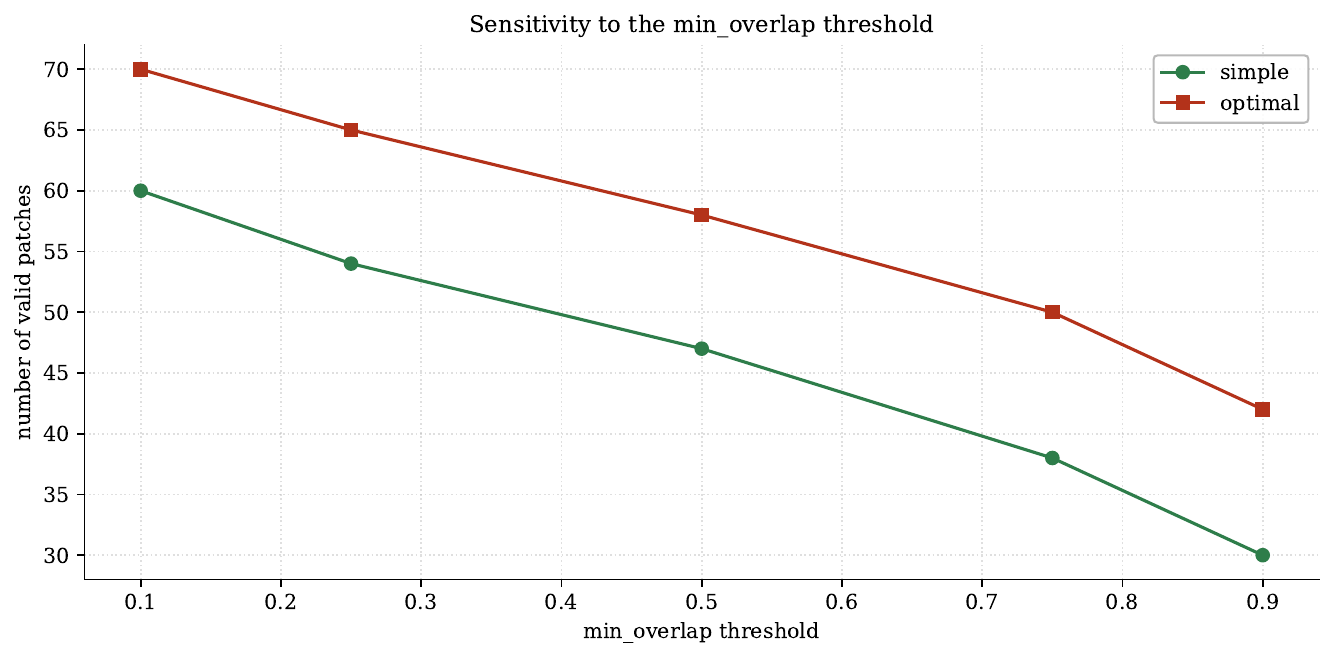}
        \caption{Minimum valid-overlap threshold}
    \end{subfigure}
    \hfill
    \begin{subfigure}[b]{0.49\linewidth}
        \centering
        \includegraphics[width=\linewidth]{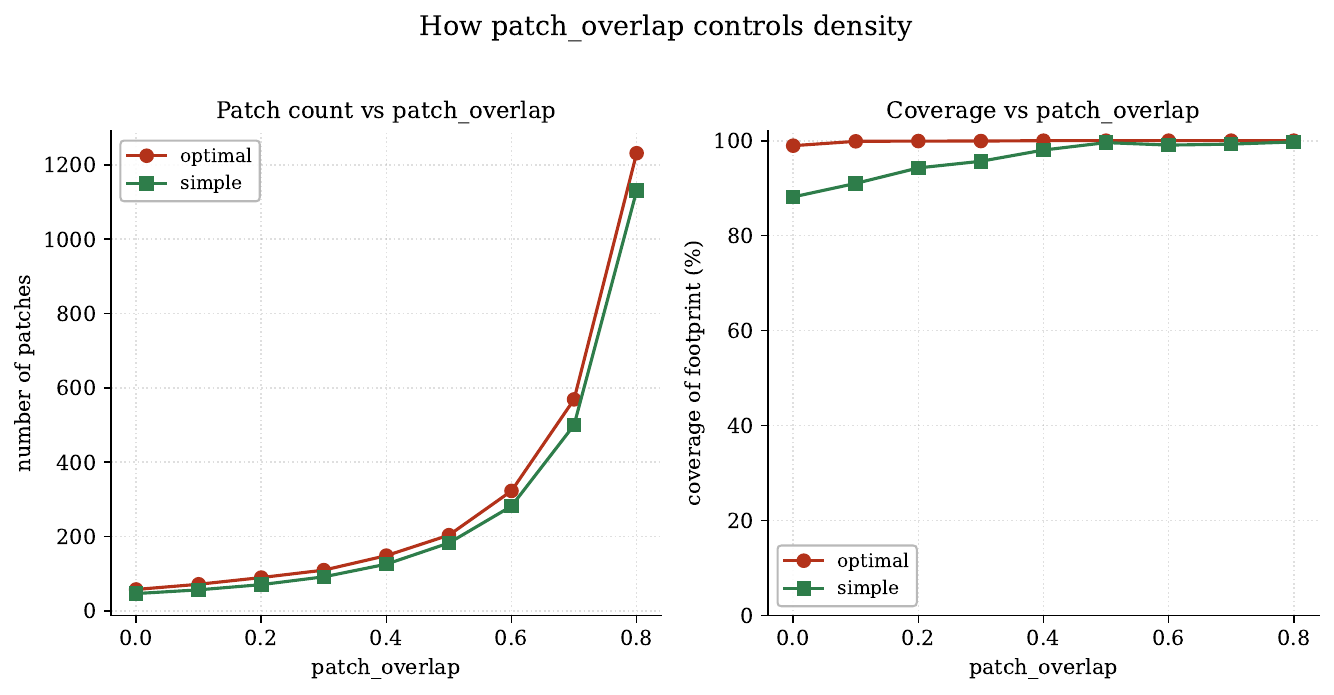}
        \caption{Patch-overlap stride}
    \end{subfigure}
    \caption{\textbf{Sampling sensitivity.} The sweeps show how patch yield and coverage respond to stricter valid-overlap requirements and different patch-overlap strides. These curves justify treating patch sampling as an explicit design choice rather than a hidden preprocessing detail.}
    \label{fig:sampling_sweeps_appendix}
\end{figure}

Training directly from JPEG2000 products is inefficient because small random windows can require decoding large internal codeblocks. The preprocessing path converts products into tiled Cloud-Optimized GeoTIFFs, enabling random patch reads that scale with the patch window rather than the full observation.

\begin{figure}[!htbp]
    \centering
    \includegraphics[height=0.8\textheight]{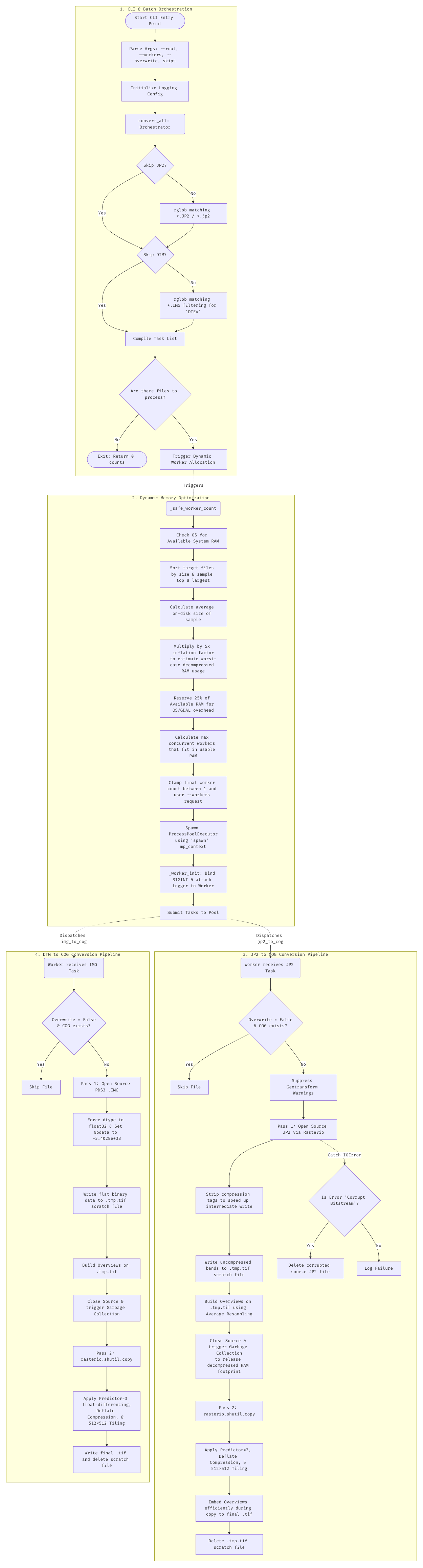}
    \caption{\textbf{Preprocessing workflow.} JPEG2000 HiRISE products are transformed into tiled, metadata-preserving assets for efficient random patch access during training.}
    \label{fig:dataset_preprocessing_appendix}
\end{figure}

\end{document}